\pdfoutput=1

\documentclass[11pt]{article}

\usepackage[final]{acl}

\usepackage{times}
\usepackage{latexsym}

\usepackage[T1]{fontenc}

\usepackage[utf8]{inputenc}

\usepackage{microtype}

\usepackage{inconsolata}
\usepackage{microtype}
\usepackage{graphicx}
\usepackage{import}
\usepackage{adjustbox}
\usepackage{layout}
\usepackage{tabularx, makecell}
\usepackage{booktabs}
\usepackage{mathrsfs}
\usepackage{amssymb} 
\usepackage{url}
\usepackage{hyperref}
\usepackage{graphicx}
\usepackage{xspace,paralist}
\usepackage{times,latexsym}
\usepackage{amsmath}
\usepackage{appendix}
\usepackage{comment} 
\usepackage{enumitem}
\usepackage{makecell}
\usepackage{multirow}
\usepackage{xcolor}
\usepackage{arydshln}
\usepackage{cleveref}
\usepackage{todonotes}
\usepackage{longtable,supertabular}
\usepackage{amssymb}
\usepackage{pifont}
\usepackage{CJKutf8}
\usepackage{subcaption}
\usepackage{ragged2e}

\newcommand{\eqnref}[1]{Eq~\eqref{#1}\xspace}

\newcommand{\appref}[1]{Appendix~\ref{#1}\xspace}

\NewDocumentCommand{\revanth}
{ mO{} }{\textcolor{blue}{\textsuperscript{\textit{Revanth}}\textsf{\textbf{\small[#1]}}}}
\title{{FAID:} \underline{{F}}ine-Grained \underline{{AI}}-Generated {T}ext \underline{{D}}etection\\ Using {M}ulti-Task {A}uxiliary and {M}ulti-Level {C}ontrastive {L}earning}

\author{Minh Ngoc Ta\textsuperscript{1,2},
Dong Cao Van\textsuperscript{1*},
Duc-Anh Hoang\textsuperscript{1*},
Minh Le-Anh\textsuperscript{1*}, 
\\\bf
Truong Nguyen\textsuperscript{1*}, 
My Anh Tran Nguyen\textsuperscript{1*}, 
Yuxia Wang\textsuperscript{2,3},
\\\bf
Preslav Nakov\textsuperscript{2}, 
Dinh Viet Sang\textsuperscript{1}
\\
\textsuperscript{1}BKAI Research Center, Hanoi University of Science and Technology \quad
\textsuperscript{2}MBZUAI\\
\textsuperscript{3}INSAIT, Sofia University "St. Kliment Ohridski"
\\
{\texttt{minh.ta@mbzuai.ac.ae, sangdv@soict.hust.edu.vn}}
}

\begin{document}
\maketitle
\def\thefootnote{*}\footnotetext{Equal contribution.}\def\thefootnote{\arabic{footnote}}

\begin{abstract}
The growing collaboration between humans and LLMs in generative tasks has introduced new challenges in distinguishing between \textit{human-written}, \textit{LLM-generated}, and \textit{human--LLM collaborative} texts. In this work, we collect a multilingual, multi-domain, multi-generator dataset \emph{FAIDSet}. We further introduce a fine-grained detection framework, \emph{FAID}, to classify text into these three categories and to identify the underlying LLM family of the generator. Unlike existing binary classifiers, FAID is built to capture both authorship and model-specific characteristics. Our method combines multi-level contrastive learning with multi-task auxiliary classification to learn subtle stylistic cues. By modeling LLM families as distinct stylistic entities, we adapt to address distributional shifts without retraining on unseen data. Our results demonstrate that FAID outperforms several baselines, particularly improving generalization accuracy across unseen domains and new LLMs, offering a potential solution to improve transparency and accountability in AI-assisted writing. Our data and code are available at \url{https://github.com/mbzuai-nlp/FAID}
\end{abstract}

\section{Introduction}
LLMs have evolved from an assistant tool to a creator or initiator, from helping polish papers to initiating proposals and drafting essays, while humans increasingly serve as optimizers and reviewers. 
In such deeply collaborative human--LLM settings, measuring human contribution becomes challenging, yet clarifying authorship is critical for accountability and transparency, particularly in educational and academic contexts~\cite{wang-etal-2025-genai}.
This work aims to identify human involvement by distinguishing the origin of a given text (in a multilingual context): (a) fully LLM-generated, (b) fully human-written, or (c) collaboratively produced.

Numerous studies have explored multilingual LLM-generated text detection, but most have focused either on binary detection, i.e.,~human vs.\ LLM~\cite{wang-etal-2024-m4, wang-etal-2024-semeval-2024, detectllm}
or on fine-grained detection limited to English~\cite{abassy-etal-2024-llm, Koike_Kaneko_Okazaki_2024, wang-etal-2023-seqxgpt, mixset}. Moreover, both struggle with generalization to unseen domains, languages, and LLM generators~\cite{wang-etal-2024-m4gt, li-etal-2024-mage}.

Our work aims to bridge this gap by (\emph{i})~collecting a multilingual, multi-domain, and multi-generator dataset, FAIDSet, for fine-grained detection, i.e.,\ identifying a text into three categories: \textit{LLM-generated}, \textit{human-written}, and \textit{human--LLM collaborative}, and (\emph{ii})~introducing a framework \emph{FAID} to improve generalization performance.

Our dataset FAIDSet\footnote{\url{https://huggingface.co/datasets/ngocminhta/FAIDSet}} focuses on the academic field, including paper abstracts, student theses, and reports, and contains generation by a variety of families of LLMs, e.g.,\ GPT, Gemini, DeepSeek, and Llama~\cite{openai2024gpt4ocard, geminiteam2024geminifamilyhighlycapable, deepseekv3-2024, grattafiori2024llama3herdmodels}.

Our detection framework, FAID, treats each LLM as a distinct author, learning specific features in the hidden space to differentiate different authors, instead of classifying based on hand-crafted stylistic features. 
We achieve this by optimizing a language encoder with multi-level author relationship (e.g.,~the stylistic similarity between texts from the same LLM family is greater than that between a human and an LLM) to capture author-specific distinguishable signals of an input text using contrastive learning, along with the task of fine-tuning a classifier to recognize the input text's origin. 
This multitask learning process forces the encoder to reorganize the hidden space so that the representations of texts by the same authors are distributed closer together than those by different authors.

In practice, given that generations produced by LLMs from the same company tend to have similar writing styles due to similar model architecture, training data, and training strategy (see~\appref{sec:analysis} for more detail), we consider each LLM family as an ``author.''
This can also help the detector acquire prior knowledge of future LLMs from the same company. 
Our experiments show that FAID consistently outperforms other baseline detectors in both in-domain and out-of-domain evaluations.
Our contributions can be summarized as follows:
\begin{itemize}
    \item We collect a new multilingual, multi-domain, multi-generator dataset for fine-grained LLM-generated text detection with 83,350 examples. 
    
    \item We propose a detection framework, FAID, to improve generalization performance in unseen domains and generators by capturing subtle stylistic features in the hidden representation.
    
    \item We show that FAID outperforms baseline detectors, particularly in unseen domains and with unseen generators. 
    Meanwhile, the nature of FAID allows us to assess the stylistic proximity of a given text to other texts in our database, each labeled with ground-truth authorship labels.
\end{itemize}

\section{Background and Related Work}

Advancements in LLMs have fundamentally reshaped the process of text production and refinement. Rather than originating every word independently, people increasingly assume the role of post-editors or reviewers, intervening after an LLM has generated an initial draft. This collaborative paradigm extends across diverse domains, including academic publishing, journalism, education, and social media, thus making hybrid human--AI authorship an emerging norm~\cite{cheng2024beyond, coauthor}. 

Such a shift calls into binary authorship detection: human vs.\ AI. Texts can be \textit{fully human-written}, \textit{fully AI-generated}, or collaborative text (e.g.,\ \textit{human-written, AI-polished}; \textit{AI-generated, human-edited}; and \textit{deeply mixed})~\cite{beemo}. Addressing these concerns requires fine-grained authorship detection, even tracing it back to a specific LLM family, to assess the extent of human contribution~\cite{hutson2025human}. 
Ensuring transparent disclosure of LLM involvement is critical to upholding research integrity and honesty.

In response to these challenges, we collect a new dataset, FAIDSet, and propose a detection framework, FAID, which generalizes well to new domains, languages, and models, achieving consistently high accuracy and reliability.

\subsection{Fine-Grained AI-generated Text Datasets}
Many prior studies have explored fine-grained AI-generated text datasets across various forms of human--AI collaboration, e.g.,~MixSet~\citep{mixset} and Beemo~\citep{beemo}. See \appref{sec:cur-dataset} for a discussion of more datasets and detailed information regarding the label space, and the coverage of domains, languages, and LLMs for each one.

However, all these studies were limited to English. 
A substantial gap remains in the availability of large-scale, fine-grained multilingual LLM-generated text detection datasets. To bridge this gap, we collect a multilingual fine-grained LLM-generated text detection dataset, which encompasses 83k texts generated by the latest LLMs and includes diverse forms of human--LLM collaborative generations. This dataset can facilitate the development of more robust and generalizable detection models that are capable of handling complex multilingual collaborative scenarios.

\subsection{Generalization of LLM-Generated Text Detection}
A recent study, M4GT-Bench, \cite{wang-etal-2024-m4gt} has highlighted a persistent challenge for both binary and fine-grained AI-generated text detection: poor generalization to unseen domains, languages, and generators. Many detection methods have shown a significant drop in performance on out-of-distribution data, underscoring the difficulty of building robust detectors for real-world scenarios and evolving LLM outputs.

Various approaches have been proposed to improve generalization. OUTFOX \cite{Koike_Kaneko_Okazaki_2024} leveraged adversarial in-context learning to dynamically generate challenging examples that enhance robustness, but still faced limitations in domain transferability and computational efficiency. LLM-DetectAIve \cite{abassy-etal-2024-llm} adopted fine-grained classification and incorporated domain-adversarial training to reduce overfitting; however, its generalization to unseen domains and generators remains limited, and its current version lacks multilingual support.

SeqXGPT \cite{wang-etal-2023-seqxgpt} focused on sentence-level detection by combining log-probabilities with convolutional and self-attention mechanisms, thereby helping capture subtle mixed-content signals and improving generalization across input styles. However, its reliance on specific model features and its limited semantic representation constrained the adaptability to new generators and domains.

Finally, the DeTeCtive framework \cite{NEURIPS2024_a117a3cd} introduced multi-level contrastive learning to better capture writing-style diversity and enhance generalizability, especially for out-of-distribution scenarios, but it primarily focused on binary classification and did not fully address \textit{human--AI collaborative} texts.

\subsection{Contrastive Learning for AI-Generated Text Detection}

Contrastive learning has been widely used to improve sentence representations by pulling semantically similar sentences closer together and pushing dissimilar ones apart.
SimCSE treats a sentence subjected to dropout noise as a semantically similar counterpart (i.e., a positive pair) and trains the encoder to minimize the distance between the original and the noise sentence. It further leverages natural language inference pairs, considering entailment pairs as positives and contradiction pairs as hard negatives~\citep{princetonsimcse}. It is then trained to maximize the separation between negative pairs in the embedding space.
DeCLUTR~\cite{giorgi2021declutrdeepcontrastivelearning} constructed positive pairs by extracting different spans from the same texts and by sampling negative pairs from different texts.


We adopt the same core idea, but reorganize the latent space by clustering \textit{human-written} texts by writing style, keeping them distant from \textit{LLM-generated} texts. Similarly to semantic textual similarity tasks, where sentence similarity ranges from 0 to 5 to reflect varying degrees of semantic overlap, we incorporate ordinal regression into our framework to model the degree of human involvement, ranging from 0 (solely LLM) to 1 (fully human).

Another work, DeTeCtive~\cite{NEURIPS2024_a117a3cd}, also leveraged contrastive learning and was used for binary detection task. 
Based on a multi-task framework, it was trained to learn style diversity using a multi-level contrastive loss, and an auxiliary task of classifying the source of a given text (human vs.\ AI) to capture distinguishable signals.

For inference, their pipeline encoded the input text as a hidden vector and used dense retrieval to match the cluster based on stylistic similarity against a database of previously indexed training features.
Additionally, instead of retraining the model on new data, they encoded the new data with the trained encoder to obtain embeddings, then added them to the feature database to augment. This largely improved the generalizability to unseen domains and new generators. 

However, this approach distinguishes only between two categories of text (\emph{human-written} vs. \emph{LLM-generated}) while overlooking the increasingly prevalent class of human--LLM collaborative texts. Our work bridges this gap to enhance generalization performance in fine-grained LLM-generated text detection.

\section{FAIDSet}
\label{sec:dataset}

We collected a new multilingual, multi-domain, and multi-generator LLM-generated text dataset, FAIDSet. It contains texts generated by LLMs, written by humans, and collaborated by both, resulting in a total of 83,350 examples; see \appref{sec:faidset-stat-analyze} for more details about FAIDSet. 

FAIDSet covers two domains: student theses and paper abstracts, where identifying authorship is critical, across two languages (Vietnamese and English). We collected students' theses from the database of Hanoi University of Science and Technology, and paper abstracts from arXiv and Vietnam Journals Online\footnote{\texttt{\url{https://vjol.info.vn/}}} (VJOL).

\textbf{Models and Label Space.}
We used the following multilingual LLM families to produce \textit{LLM} and \textit{human--LLM collaborative} texts: GPT-4/4o, Llama-3.x, Gemini 2.x, and Deepseek V3/R1.
Regarding the \textit{human--LLM collaborative} text, we include \textit{LLM-polished}, \textit{LLM-continued}, and \textit{LLM-paraphrased}, where the models are requested to polish or paraphrase inputs while ensuring the accuracy of any figures and statistics.

\textbf{Diverse Prompt Strategies.}
We generated data with diverse tones and contexts while ensuring content accuracy. Depending on the data source and context, we crafted prompts to create varied outputs suitable for different real-world scenarios. We generated responses with different tones using prompts such as \textit{``You are an IT student...''} and \textit{``...who are very familiar with abstract writing...''}. See Appendix \ref{sec:diverse-prompt} for the full list.

While FAIDSet does not capture the full natural diversity of in-the-wild LLMs' outputs, the controlled setup enables us to systematically model stylistic and collaborative signals across multiple languages and families. This makes FAIDSet both a reproducible training resource and a benchmarking corpus: it provides reliable supervision for developing detectors while also serving as a testbed for assessing generalization to more diverse, unseen scenarios.

\textbf{Quality Control.} 
To avoid bad machine-generated texts, which can introduce remarkably distinguishable signals, we performed quality control by randomly sampling 10--20 instances for each domain, source, and LLM generator. Manual inspection focused on fluency, coherence, and factual plausibility. Overall, the generated texts demonstrated high linguistic quality, with most outputs being fluent and logically reasonable. Nevertheless, occasional issues such as repetitive phrasing, incomplete reasoning, or overuse of formal expressions were observed. In those cases, we adjusted the prompts (e.g., by specifying desired length or style) or refined generation parameters to improve diversity and logical consistency. After these adjustments, the quality across generators and domains was found to be stable and satisfactory.
\section{Methodology}

\paragraph{Task Definition}
Our task is a three-class classification problem: \textit{human-written} vs. \textit{LLM-generated} vs. \textit{human--LLM collaborative} text detection. 

The \textit{human--LLM collaborative} category involves a range of interactions between humans and LLM systems, such as (a)~\textit{human-written, LLM-polished}, (b)~\textit{human-initiated, LLM-continued}, (c)~\textit{human-written, LLM-paraphrased}, etc.
Given the growing variety and complexity (e.g.,\ \textit{deeply mixed} text) of collaborative patterns, this is not exhaustive. Instead, we consolidate all forms of human--LLM collaboration into a single label to maintain practical simplicity and model generalizability. This reflects real-world usage, where using LLM tools to enhance clarity or expression is increasingly common and often ethically acceptable.

Our analysis of the dataset revealed that the LLM models within the same family tend to have similar writing style and text distributions, due to their shared training data and architecture (see \appref{sec:analysis}). Thus, we consider each model's family to be an ``author'' with a unique writing style.

\subsection{Framework Overview}

\begin{figure*}
    \centering
    \includegraphics[width=0.9\linewidth]{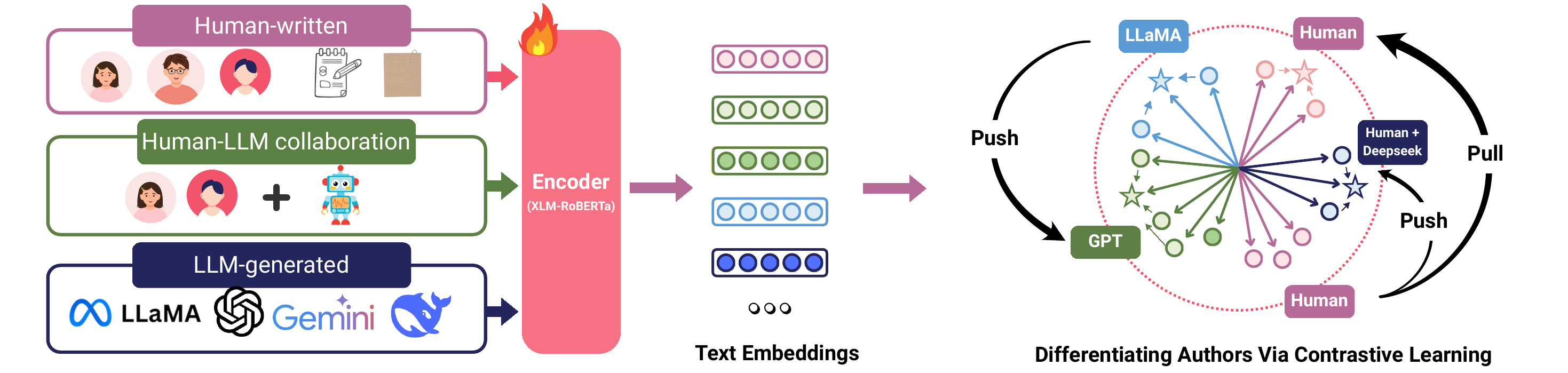}
    \caption{Training architecture. Leveraging multi-level contrastive learning loss, we fine-tune a language model (we select XLM-RoBERTa~\cite{ethroberta}, see Appendix \ref{sec:model-select}) based on the human, human--LLM and LLM-generated texts, to force the model to reorganize the hidden space, pulling the embeddings within the same author families closer, and pushing the embeddings from different authors farther. We train an encoder that can represent text with distinguishable signals to discern the authorship of text.} 
    \label{fig:train}
\end{figure*}

 Here, we shift the focus to the \textit{detector}, an encoder-based model that forms the core of \textsc{FAID}. The detector encodes each text into a high-dimensional embedding space to quantify cross-source similarity, enabling the study of both intra- and inter-family relationships. It is trained to capture \textit{multi-level similarities between authors} by learning a representation space where closely related sources (e.g., LLMs from the same family) form tighter clusters, while dissimilar ones (e.g., human vs.\ LLM) are pushed farther apart. 

Let $\mathrm{S}_c$ be cosine similarity, $\phi(\cdot)$ be the encoder function, and $P_i$, $P_j$, ($1 \leq i \leq j < 5$) be the distributions of different text sources. We aim for the model to encode representations that satisfy the following constraint:
\begin{equation}
    \begin{split}
        \mathbb{E}_{x \in P_i, y \in P_j} & \left[ \mathrm{S}_c \left( \phi(x), \phi(y) \right) \right] \\
        & \geq \mathbb{E}_{x \in P_i, y \in P_{j+1}}  \left[ \mathrm{S}_c \left( \phi(x), \phi(y) \right) \right]
    \end{split}
    \label{eqn:relation} 
\end{equation}

\noindent where $P_1$ corresponds to the distribution generated by a particular LLM family, $P_2$ is the distribution generated by any LLM, $P_3$ is the distribution of \textit{collaborative} text generated by human and a LLM family of $P_1$, $P_4$ is the distribution of \textit{collaborative} text generated by humans and any LLM families, and $P_5$ is the distribution of \textit{human-written} text.


To clarify the rationale for configuring FAID to expect that the similarity of a text $x$ (from lower-level distributions $P_1$ or $P_2$) with samples from $P_3$ is generally greater than or equal to its similarity with samples from $P_4$, consider the following: 

\begin{itemize}
    \item If $x \in P_1$ ($x$ is generated by a \textit{particular} LLM): Let $y_{\textit{LHS}}$ be drawn from $P_3$ and $y_{\textit{RHS}}$ be drawn from $P_4$. Naturally, the similarity $\mathrm{S}_c(x, y_{\textit{LHS}})$ is greater than $\mathrm{S}_c(x, y_{\textit{RHS}})$. This is because $P_3$ contains texts that share a direct LLM family origin with $x$.
    \item If $x \in P_2$ ($x$ is generated by \textit{any} LLM): Here, with $y_{\textit{LHS}}$ from $P_3$ and $y_{\textit{RHS}}$ from $P_4$ as defined above, the similarity $\mathrm{S}_c(x, y_{\textit{LHS}})$ is generally expected to be equal to $\mathrm{S}_c(x, y_{\textit{RHS}})$. Since $x$ can originate from any LLM, it does not inherently possess a stronger connection to the specific LLM family in $P_3$ than to the broader human-LLM collaborations represented in $P_4$.
\end{itemize}

This configuration aims to ensure that closeness in distribution corresponds to higher similarity after encoding, thus encouraging the model to discern fine-grained multi-level relations.

\begin{figure*}
    \centering
    \includegraphics[width=0.9\linewidth]{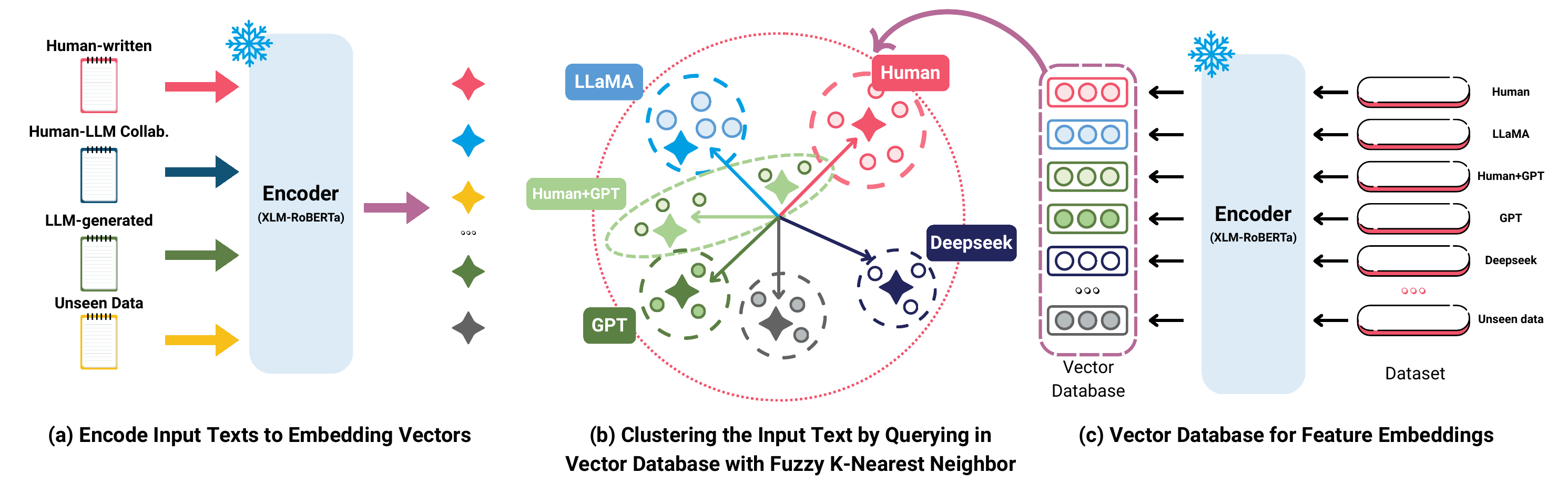}
    \caption{Inference architecture: (a)~embed the input text into embedding vector using the fine-tuned encoder, (b)~use Fuzzy kNN to cluster, retrieving which cluster the input text belongs to (see more in Appendix~\ref{sec:ablation-study}), (c)~the stored vector database $\mathcal{VD}$ was created by saving all embeddings of texts in training and validation sets using the fine-tuned encoder. If the input text is unseen, we embed it and save it into a temporary vector database $\mathcal{VD}'$, enhancing the generalization of the detector.}
    \label{fig:infer}
\end{figure*}

\subsection{Multi-level Contrastive Learning}
Given a dataset with $N$ examples, 
each example is a text unit (paragraph/segment). The $i^{th}$ record is denoted as $T_i$. Per record, we assigned three-level labels indicating its source:
\begin{itemize}
    \item $x_i \in \{0,1\}$: if $T_i$ is a fully LLM-generated text, $x_i=0$, otherwise $x_i=1$; 
    \item $y_i \in \{0,1\}$: if $T_i$ is a fully human-written text, $y_i=0$, otherwise $y_i=1$;
    \item $z_i$: an indicator of a specific LLM family.
\end{itemize}

The encoder $\phi(\cdot)$ represents the text $T_i$ in a $d$-dimensional vector space $\mathbb{R}^d$. Then we calculate the cosine similarity between two texts $T_i$ and $T_j$, denoted by: $\sigma(i,j) = \mathrm{S}_c\left( \phi(T_i), \phi(T_j) \right)$.

For \textit{LLM-generated} text, the similarity between $T_i$ and another \textit{LLM-generated} text $T_j$ is greater than that with a \textit{human-written} or \textit{collaborative} text $T_k$:
\begin{equation}
    \sigma(i,j) > \sigma(i,k), \forall x_i = 0, x_i = x_j, x_k = 1
    \label{eqn:rel1}
\end{equation}

If $x_i=0$, then the text is fully \textit{LLM-generated}. For this case, we do not consider $y_i$ since \textit{LLM-generated} text is considered as non-LLM-collaboration. We can imply that the similarity between two texts written by the same LLM is higher than that of two LLM families. Hence: 
\begin{equation}
    \sigma(i,j) > \sigma(i,k), \forall x_i = 0, z_i = z_j, z_i \neq z_k
    \label{eqn:rel2}
\end{equation}

The reverse condition is also true. We can conclude that:
\begin{equation}
    \sigma(i,j) > \sigma(i,k), \forall x_i = 1, y_i = y_j, y_i \neq y_k
    \label{eqn:rel3}
\end{equation}

For cases where all samples are \textit{human--LLM collaborative}, two texts created by the same LLM family tend to be more similar than such that involve contributions from different LLM families. That is:
\begin{equation}
    \sigma(i,j) > \sigma(i,k), \forall x_i = 1, y_{i,j,k} = 1, z_i = z_j \neq z_k
    \label{eqn:rel4}
\end{equation}

Combining all, the text representation is learned with the following constraints:
\begin{equation}
    \resizebox{0.88\columnwidth}{!}{
    $\begin{cases}
        \sigma(i,j) > \sigma(i,k), \forall x_i = 0, x_i = x_j \neq x_k;\\
        \sigma(i,j) > \sigma(i,k), \forall x_i = 0, z_i = z_j \neq z_k;\\
        \sigma(i,j) > \sigma(i,k), \forall x_i = 1, x_i = x_j \neq x_k;\\
        \sigma(i,j) > \sigma(i,k), \forall x_i = 1, y_i = y_j \neq y_k;\\
        \begin{split}
            \sigma(i,j) > \sigma(i,k), \forall & x_i = 1, y_{i,j,k} = 1, \\
            & z_i = z_j, z_i \neq z_k\\ 
        \end{split}
    \end{cases}$
    }
    \label{eqn:constrain}
\end{equation}

To enforce the similarity constraints outlined in \eqnref{eqn:constrain}, we build upon the SimCLR framework \cite{pmlr-v119-chen20j} and introduce a strategy for defining both positive and negative sample pairs, which forms the basis of our contrastive learning loss. Departing from traditional contrastive losses that rely on a single positive sample, our approach considers a group of positive instances that satisfy specific criteria. The similarity between the anchor and the positive samples is computed as the average similarity across this entire positive set. For negative samples, we follow the methodology used in SimCLR. The resulting contrastive loss, expressed in \eqnref{eqn:simclr}, involves $q$ as the anchor sample, $K^+$ as the positive sample set, $K^-$ as the negative sample set, $\tau$ as the temperature parameter, and $N_{K^+}$ as the number of positive samples.

\begin{equation}
    \resizebox{0.88\columnwidth}{!}{
    $\mathcal{L}_q = -\log \frac{
        \exp\left( \sum_{k \in K^+} \frac{S(q, k)}{\tau} / N_{K^+} \right)
    }{
        \exp\left( \sum_{k \in K^+} \frac{S(q, k)}{\tau} / N_{K^+} \right)
        + \sum_{k \in K^-} \exp\left( \frac{S(q, k)}{\tau} \right)
    }
    $}
    \label{eqn:simclr}
\end{equation}

Different constraints yield different sets of positive and negative samples. Based on these sets, contrastive losses are computed at multiple levels. As we declared in \eqnref{eqn:simclr}, each inequality in \eqnref{eqn:constrain} is denoted as $\mathcal{L}_{q_i,\varepsilon}$ where $\varepsilon = \overline{1,5}$ respectively. To form \eqnref{eqn:lmcl}, we need to add the coefficients $\alpha, \beta,\gamma,\delta$, and $\zeta$ to maintain the balance of multi-level relations.

\begin{equation}
    \begin{split}
        \mathcal{L}_{mcl} = & \sum\limits_{i=1}^{N} \left[ x_i \left( \alpha \mathcal{L}_{q_i,1} + \beta \mathcal{L}_{q_i,2} \right) + \right. \\
        & \left. \left( 1-x_i\right) \left( \gamma \mathcal{L}_{q_i,3} + \delta \mathcal{L}_{q_i,4} + \zeta \mathcal{L}_{q_i,5} \right) \right]
    \end{split}
    \label{eqn:lmcl}
\end{equation}

Due to the last inequality in \eqnref{eqn:constrain} only specifying a case ($y_{i,j,k}=1$), and the other cases considering both values for $y$, we have $\zeta = 2\gamma = 2\delta$. Also, to maintain the equilibrium, we need to keep $\alpha + \beta = \gamma + \delta + \zeta$. We set $\gamma = \delta = 1$, then $\zeta = 2, \alpha = \beta = 2$.
This encourages the model to capture subtle and detailed features from different sources. As a result, it becomes more adept at recognizing variations in writing styles. This capability enhances accuracy and strengthens generalizability when detecting \textit{LLM-generated} text.

\subsection{Multi-Task Auxiliary Learning}
Multi-task learning \cite{caruana1997multitask} allows a model to learn several tasks concurrently by sharing relevant information across them. This joint learning process helps the model develop more general and distinctive features. Therefore, it improves the model's generalizability to new data. Building on the previously described contrastive learning framework, we extend the encoder by attaching an MLP classifier at its output layer.

This classifier performs binary classification, determining whether a given text was written by a human or an LLM. Let the probability of $i^{th}$ sample with label $x_i = 0$ be $p_i$. To train this component, we apply a cross-entropy loss function, denoted as $\mathcal{L}_{ce}$, and defined as follows:
\begin{equation}
    \mathcal{L}_{ce} = -\dfrac{1}{N}\sum\limits_{i=1}^{N} x_i \log p_i + \left( 1-x_i \right) \log \left( 1-p_i \right)
    \label{eqn:lce}
\end{equation}

Therefore, the overall loss is computed as:
\begin{equation}
    \mathcal{L} = \mathcal{L}_{ce} + \mathcal{L}_{mcl}
    \label{eqn:ovrloss}
\end{equation}

\subsection{Handling Unseen Data without Retraining}
Unseen data, whether from an unseen domain or an unfamiliar generator, remains a significant challenge even for state-of-the-art LLM-generated text detection methods, as LLMs continue to improve. We tried to use the model classifier on its own, but we ended up using a vector database along with Fuzzy k-Nearest Neighbors, as illustrated in Figure~\ref{fig:infer}. The results are given in Appendix \ref{sec:ablation-study}. Specifically, when dealing with unseen data, we use our model to embed these texts and add them to our existing vector database. Through careful parameter tuning, this approach enables our system to effectively handle newly encountered unseen data without retraining.
\section{Experiments} 
\label{sec:exp}
\begin{table*}[t!]
    \centering
    \small
    \begin{tabular}{llcccccc}
    \toprule
    \textbf{Dataset} & \textbf{Detector} & \textbf{Accuracy $\uparrow$} & \textbf{Precision $\uparrow$} & \textbf{Recall $\uparrow$} & \textbf{F1-macro $\uparrow$} & \textbf{MSE $\downarrow$} & \textbf{MAE $\downarrow$} \\
    \midrule
    \multirow{4}{*}{FAIDSet} 
    & LLM-DetectAIve & \underline{94.34} & 94.45 & \underline{93.79} & \underline{94.10} & \underline{0.1888} & 0.1107\\
    & T5-Sentinel & 93.31 & \underline{94.92} & 93.10 & 93.15 & 0.2104 & \underline{0.1101} \\
    & SeqXGPT &  85.77&  85.49&  86.02&  84.69&  0.5593& 0.2844\\
    & FAID & \textbf{95.58} & \textbf{95.78} & \textbf{95.33} & \textbf{95.54} & \textbf{0.1719} & \textbf{0.0875}  \\
    \midrule
    \multirow{4}{*}{LLM-DetectAIve} 
    & LLM-DetectAIve & \underline{95.71} & \textbf{95.78} & \textbf{95.72} & \textbf{95.71} & \underline{0.1606} & \underline{0.1314} \\ 
    & T5-Sentinel & 94.77  & 94.70 & \underline{92.60} & \underline{93.60} & 0.1663 &  0.1503\\
    & SeqXGPT &  81.48&  78.72&  74.91&  76.71&  0.3141& 0.2255\\
    & FAID & \textbf{96.99} & \underline{95.29} & 88.14 & 91.58 & \textbf{0.1561} & \textbf{0.0754}  \\
    \midrule
    \multirow{4}{*}{HART} 
    & LLM-DetectAIve & \underline{94.39} & \underline{94.25} & \underline{94.33} & \underline{94.29} & \textbf{0.3244} & \underline{0.1789}\\
    & T5-Sentinel & 86.68 & 87.25 & 87.69 & 87.38  & \underline{0.4339} & 0.2334 \\
    & SeqXGPT &  63.12 & 64.01 & 65.27 & 64.05 & 1.0057 & 0.5982 \\
    & FAID & \textbf{96.73} & \textbf{97.61} & \textbf{98.05} & \textbf{97.80}  & 0.4631 & \textbf{0.1806} \\
    \bottomrule
    \end{tabular}
    \caption{Performance with three labels. The best results are in \textbf{bold} and the second best are \underline{underlined}.}
    \label{tab:3labels-acc}
\end{table*}

In this section, we describe the datasets and baselines we used, followed by two experiments evaluating FAID: (\emph{i})~classify a text as \textit{human}, \textit{LLM}, and \textit{human--LLM}, and (\emph{ii})~identify specific generators.

\subsection{Datasets}
In addition to FAIDSet, for in-domain evaluation, we used two additional datasets:

\textbf{LLM-DetectAIve \cite{abassy-etal-2024-llm}} encompasses various domains, including arXiv, Wikihow, Wikipedia, Reddit, student essays, and peer reviews. We augmented the original labels \textit{human-written} and \textit{machine-generated} using multiple LLMs to create a 485,405-example dataset with two new labels: (\emph{i})~\textit{machine-written then machine-humanized}, and (\emph{ii})~\textit{human-written then machine-polished}.

\textbf{HART \cite{bao2025decouplingcontentexpressiontwodimensional}} has 21,500 examples, including student essays, arXiv abstracts, story writing, and news articles. It covers four categories: \textit{human-written}, \textit{AI-refined}, \textit{AI-generated}, and \textit{humanized AI-generated} texts. The authors further expanded the dataset by creating additional instances with unbalanced label distributions.

To evaluate the generalizability of FAID on unseen scenarios, we collected the following data:

\textbf{Unseen domain:} We created a dataset consisting of 150 IELTS essays from Kaggle\footnote{\url{https://www.kaggle.com/datasets/mazlumi/ielts-writing-scored-essays-dataset}}, where all texts are \textit{human-written}. We used these essays to generate \textit{human--LLM collaborative} and \textit{LLM-generated} texts with the same models with FAIDSet.

\textbf{Unseen generators:} We selected 150 human-written abstracts from the FAIDSet test set and generated data for the remaining labels using three new LLM families: Qwen, Mistral, and Gemma.

\textbf{Unseen domain \& generators:} Based on the \textit{human-written} IELTS essays above, we used the same LLM families as for the unseen generator test set to generate data for the LLM and the human--LLM labels.

\begin{table*}[t!]
    \centering
    \small
    \resizebox{\textwidth}{!}{
    \begin{tabular}{llcccccc}
    \toprule
    \textbf{Dataset} & \textbf{Detector} & \textbf{Accuracy $\uparrow$} & \textbf{Precision $\uparrow$} & \textbf{Recall $\uparrow$} & \textbf{F1-macro $\uparrow$} & \textbf{MSE $\downarrow$} & \textbf{MAE $\downarrow$} \\
    \midrule
    \multirow{4}{*}{Unseen domain}
    & LLM-DetectAIve & 52.83 & 47.31 & 64.62 & 53.28 & 0.4733 & 0.4722\\
    & T5-Sentinel & \underline{55.56} & \underline{49.54} & \underline{66.67} & \underline{55.34}  & \textbf{0.4444} & \textbf{0.4444} \\
    & SeqXGPT & 40.60 & 43.81 & 31.87 & 36.72 & 0.8021 & 0.7028 \\
    & FAID & \textbf{62.78} & \textbf{70.73} & \textbf{71.77} & \textbf{69.46} & \underline{0.4514} & \underline{0.4486} \\
    \midrule
    \multirow{4}{*}{\makecell[l]{Unseen generators}}
    & LLM-DetectAIve & 75.71 & 73.25 & 75.63 & 74.30 & 0.3714 & 0.2957\\
    & T5-Sentinel & \underline{85.95} & \underline{85.77} & \underline{84.59} & \underline{85.16} & \underline{0.3648} & \underline{0.2419} \\
    & SeqXGPT & 72.04 & 60.33 & 48.94 & 54.12 & 0.4590 & 0.3380 \\
    & FAID & \textbf{93.31} & \textbf{92.40} & \textbf{94.44} & \textbf{93.25} & \textbf{0.1691} & \textbf{0.1167} \\
    \midrule
    \multirow{4}{*}{\makecell[l]{Unseen domains and\\ Unseen generators}} 
    & LLM-DetectAIve & \underline{62.93} & \underline{66.74} & \underline{71.17} & \underline{61.97} & 0.4479 & \underline{0.3964} \\
    & T5-Sentinel & 57.07 & 49.82 & 66.61 & 55.45  & \underline{0.4314} & 0.4300 \\
    & SeqXGPT & 40.71 & 47.95 & 35.21 & 40.09 & 0.8753 & 0.7086 \\
    & FAID & \textbf{66.55} & \textbf{74.44} & \textbf{73.57} & \textbf{72.58} & \textbf{0.3939} & \textbf{0.3167} \\
    \bottomrule
    \end{tabular}
   }
    \caption{Performance with three labels with unseen data. We use the detector trained on FAIDSet and evaluate on the unseen datasets. The best results are in \textbf{bold} and the second best are \underline{underlined}.}
    \label{tab:3labels-acc-unseen}
\end{table*}

\begin{table*}[t!]
    \centering
    \small
    \begin{tabular}{llcccc}
    \toprule
    \textbf{Dataset} & \textbf{Detector} & \textbf{Accuracy $\uparrow$} & \textbf{Precision $\uparrow$} & \textbf{Recall $\uparrow$} & \textbf{F1-macro $\uparrow$} \\
    \midrule
    \multirow{4}{*}{FAIDSet} 
    & LLM-DetectAIve & \underline{75.96} & 76.97 & 76.90 & 76.53 \\
    & T5-Sentinel & 75.68 & \underline{79.85} & \underline{78.40} & \underline{78.37} \\
    & SeqXGPT & 69.41 & 68.02 & 64.20 & 66.03 \\
    & FAID & \textbf{79.64} & \textbf{83.28} & \textbf{83.52} & \textbf{83.27} \\
    \midrule
    \multirow{4}{*}{LLM-DetectAIve} 
    & LLM-DetectAIve & \underline{}{90.49} & \textbf{90.64} & \underline{83.52} & \underline{86.93} \\
    & T5-Sentinel & 81.54  & 81.37 & 80.09  & 81.05 \\
    & SeqXGPT & 87.12 & 78.90 & 76.08 & 74.41 \\
    & FAID & \textbf{90.89} & \underline{88.17} & \textbf{86.72} & \textbf{87.37} \\
    \midrule
    \multirow{4}{*}{HART} 
    & LLM-DetectAIve & \underline{89.00} & \underline{87.87} & \textbf{86.74} & \textbf{87.15} \\
    & T5-Sentinel & 78.52 & 77.13 & 78.34 & 77.59 \\
    & SeqXGPT & 64.70 & 55.82 & 45.40 & 50.75 \\
    & FAID & \textbf{89.96} & \textbf{91.57} & \underline{86.48} & \underline{86.67} \\
    \bottomrule
    \end{tabular}
    \caption{Accuracy of identifying generators: human, GPT, Gemini, Deepseek, and Llama. The best is in \textbf{bold}.}
    \label{tab:model-acc}
\end{table*}

\subsection{Baselines}

\paragraph{LLM-DetectAIve:}
We adapted the method of \citet{abassy-etal-2024-llm} by fine-tuning a \texttt{roberta-base} sequence classification model. We tokenized the input texts using the RoBERTa tokenizer with a maximum sequence length of 256, and we trained the model for three-class detection.

\paragraph{T5-Sentinel:} We adapted the T5-Sentinel framework introduced by \citet{chen-etal-2023-token} for our three-class setup. Following the original configuration, we trained using the AdamW optimizer (batch size 128, learning rate $1 \times 10^{-4}$, weight decay $5 \times 10^{-5}$). This allows direct comparison with prior T5-based detectors under our experimental conditions.

\paragraph{SeqXGPT:}
We adopted the method of \citet{wang-etal-2023-seqxgpt}, who model token-level likelihood patterns from LLM tokenizers for sentence-level detection: we updated the tokenizer models to align with our dataset's label space. This adjustment ensures that the extracted token-level log-probability features better reflect the model types in our data.

\subsection{Human-Only, LLM-Only, or Human-LLM?}
Tables~\ref{tab:3labels-acc} and \ref{tab:3labels-acc-unseen} show the performance of FAID and three baselines in three evaluation settings.

As shown in Table~\ref{tab:3labels-acc}, FAID consistently achieves the best accuracy in (\emph{i})~in-domain and known generators, (\emph{ii})~unseen domains, (\emph{iii})~unseen generators, and (\emph{iv})~unseen domain \& generators settings. It is followed by LLM-DetectAIve for (\emph{i}) and (\emph{iv}), and by T5-Sentinel for (\emph{ii}) and (\emph{iii}).
Despite being designed to extract sequence-level features, SeqXGPT struggles with texts from advanced models, whose coherent, human-like writing styles reduce the detectable distinctions.

FAID further improves generalization performance over unseen domains and generators compared to others as illustrated in Table~\ref{tab:3labels-acc-unseen}. 
We can see that generalizing to (\emph{ii}) unseen domains and (\emph{iv}) unseen domain \& generator remains challenging, with accuracy of 62.78\% and 66.55\%, respectively. 
These results suggest that FAID is an effective method for addressing the multilingual fine-grained LLM-generated text detection task, improving performance by leveraging multi-level contrastive learning to capture generalizable stylistic differences tied to LLM families, rather than overfitting to surface-level artifacts.

\subsection{Identifying Different Generators}
The goal of FAID is not only to detect whether AI was used to produce the target text, but also to identify the specific LLM family, treating the families as distinct authors. As shown in Table \ref{tab:model-acc}, FAID consistently achieves higher performance compared to other baselines in almost all evaluation measures.

LLM-DetectAIve achieves results comparable to FAID on the test set of its dataset, except that its precision is slightly lower. 
FAID's high performance when dealing with text from diverse known generators in these datasets indicates that it learned unique writing patterns and features of different generators by leveraging multi-level contrastive learning.

\begin{table}[t!]
    \centering
    \resizebox{\linewidth}{!}{
    \begin{tabular}{lcccc}
    \toprule
    \textbf{Language} & \textbf{Accuracy} & \textbf{Precision} & \textbf{Recall} & \textbf{F1-macro} \\
    \midrule
    English    & 96.41 & 96.02 & 95.59 & 95.77 \\
    Vietnamese & 94.42 & 95.60 & 94.22 & 94.76 \\
    \bottomrule
    \end{tabular}
    }
    \caption{Language-wise performance on FAIDSet.}
    \label{tab:language_wise_results}
    \vspace{-5pt}
\end{table}

\subsection{Language-wise Performance}

Table~\ref{tab:language_wise_results} presents the classification performance of FAID on each language of FAIDSet. The results show that FAID achieves consistently strong performance across both subsets. While performance on English is slightly higher, FAID maintains competitive accuracy and F1-score on Vietnamese, indicating that the model does not rely solely on high-resource language features.

\subsection{Generalizability to Unseen Human--LLM type of Collaboration}
\label{sec:gen-unseencolab}
To assess FAID’s robustness to unseen collaborative writing patterns between humans and LLMs, we conducted an additional experiment focusing on hybrid authorship styles that were not present in the original training dataset.

We conducted manual quality control to enhance the reliability of our dataset and to ensure the robustness of the FAID model. A team of five annotators, all IT-majored, fluent in English, and aged 18--25, participated in the annotation process. Each annotator was assigned approximately 80 samples, covering all collaborative writing styles, for a total of 400 human-reviewed instances across various collaboration modes.

During this manual revision stage, the annotators followed two key quality control principles:
\begin{enumerate}
    \item Ensuring logical and informational consistency between the outputs and the original texts, where the output length was controlled to be 70--150\% of the source length.
    \item Improving quality through spelling correction, synonym replacement, and word refinement to ensure natural fluency and stylistic coherence.
\end{enumerate}

FAID was evaluated on this manually curated dataset without further fine-tuning. Despite these samples representing an unseen distribution, the model achieved a strong overall accuracy of 84.8\%, precision of 82.8\%, and recall of 85.0\% across all collaboration categories. This suggests a strong generalization capability beyond the data distributions encountered during model development. It also highlights FAID's sensitivity to fine-grained stylistic blending, where human revision only partially conceals the generative footprint of LLMs.

\subsection{Generalizability to Real-World Scenarios}
To further assess FAID's generalizability beyond the controlled benchmark setting, we conducted an additional user study simulating realistic academic and professional writing scenarios. The goal was to evaluate whether FAID can maintain detection accuracy when applied to authentic, unconstrained text produced through real interactions with LLMs.

A group of five volunteers with diverse academic backgrounds was instructed to engage with four popular AI systems: ChatGPT, Gemini, DeepSeek, and Llama 3.1 to generate text resembling authentic student or professional writing.

The interactions were open-ended but were guided by three types of collaboration commonly encountered in two real-world use cases: writing a paper summary (as an abstract) and writing a passage for their own graduation thesis.
Each participant was asked to generate five outputs per model, yielding a total of 200 real-world text samples. To ensure authenticity and diversity, the participants were encouraged to adjust the prompts iteratively, to include follow-up clarifications, and to perform light editing, mimicking realistic human--LLM co-writing behavior.

FAID achieved strong performance, with overall accuracy of 88.5\%, precision of 85.9\%, and recall of 89.7\%, indicating strong generalizability to unseen real-world scenarios despite being trained only in an in-domain setting.




\section{Conclusion and Future Work}
We presented FAIDSet, a multi-domain, multilingual fine-grained LLM-generated text detection dataset comprising 83k examples, and FAID, a framework designed to distinguish between human-written, LLM-generated, and especially, human--LLM collaborative texts in practice.

FAID integrates multi-level contrastive learning and multi-task auxiliary objectives, treating LLM families as stylistic ``authors'', which enables it to capture subtle linguistic and stylistic cues that generalize effectively across domains and evolving generative systems. Moreover, its Fuzzy k-Nearest Neighbors-based inference and training-free incremental adaptation contribute to strong robustness and adaptability to unseen data.

Our experiments demonstrated that FAID consistently outperforms competitive baselines across multiple datasets and settings. Its ability to detect nuanced collaborative writing and to adapt to emerging generative models highlights its potential for real-world deployment.

In future work, we plan to extend FAIDSet to cover more languages, generators, and domains, particularly low-resource languages and informal genres such as social media and student writing, to further enhance cross-lingual and domain robustness. We further plan to incorporate adversarially LLM-generated texts as well as more complex forms of human--LLM collaboration in order to better capture the evolving dynamics of AI-assisted text creation.
\section*{Limitations}
While FAID demonstrates strong performance and generalization across various domains and LLMs, several limitations remain. First, although our dataset is multilingual and multi-domain, it remains limited in low-resource languages and niche writing domains, which may affect performance in those contexts. FAIDSet is synthetic by construction. The controlled generation enables causal-style analysis, but it under-represents the messy, tool-chain-specific edits seen in the wild. We mitigate this with diverse prompt paraphrases, manual spot-checks, and generalization tests to held-out generators. Second, our framework is based on the observation that texts produced by LLMs from the same family share similar stylistic features. However, this may break down when a single text is influenced by multiple LLMs, e.g.,~ when a human uses different models for drafting, rewriting, and polishing. In such cases, the resulting style may blend traits from multiple LLMs, making it more difficult to attribute authorship to a single LLM family or clearly distinguish collaboration boundaries.

\section*{Ethics and Broader Impact}
\paragraph{Data Collection and Licenses}
A primary ethical consideration is the data license. We reused existing datasets for our research: LLM-DetectAIve, HART, and IELTS Writing, which have been publicly released with clear licenses and well-documented terms of use. We adhere to the intended usage of these datasets.

\paragraph{Security Implications.}
FAIDSet streamlines both the creation and the rigorous testing of FAID. By spotting LLM-generated material, FAID helps preserve academic integrity, flag potential misconduct, and protect the genuine contributions of authors. More broadly, it supports efforts to prevent the misuse of generative technologies, such as credential falsification. Detecting LLM-generated content across different languages can be tricky, due to the language's grammar and style. By enabling robust, multilingual, and multi-generator detection with accurate results, FAIDSet empowers people everywhere, especially in academic scenarios, to deploy AI responsibly. At the same time, it fosters critical digital literacy, giving everyone a clear understanding of both the strengths and the limits of generative AI.

\paragraph{Responsible Use of AI-generated Text Detection.}
FAID is designed to enhance transparency in AI-assisted writing by enabling the fine-grained detection of AI involvement in text generation. While this has clear benefits for academic integrity and content provenance, we acknowledge the potential for misuse. For instance, such tools could be used to unfairly penalize individuals in educational or professional settings based on incorrect or biased predictions. To mitigate this, we stress that FAID is not intended for high-stakes decision-making without human oversight.

\paragraph{Bias and Fairness.}
AI-generated text detection systems may inadvertently encode or amplify biases present in the training data. FAIDSet has been carefully constructed to include diverse domains and languages to reduce such biases. Nonetheless, we encourage ongoing auditing and benchmarking of fairness across populations and writing styles, and welcome community feedback for further improvements.

\paragraph{Transparency and Reproducibility.} We promote open research and community contributions, and thus we publish our code and data.

\bibliography{ref}
\bibliographystyle{acl_natbib}

\appendix
\clearpage
\section*{Appendix}
\appendix
\section{Some Current Datasets on AI-generated Texts}
\label{sec:cur-dataset}

We review existing datasets for AI-generated text detection and summarize their key characteristics in Table~\ref{tab:cur-dataset}. While many of these datasets provide fine-grained labels and cover multiple text generators, all are monolingual and limited to English. This limitation highlights a critical gap in current resources and motivates the construction of a new multilingual, multi-domain, and multi-generator dataset for AI-generated text detection, which is essential for developing methods that generalize to real-world, cross-lingual scenarios.


\begin{table*}
    \centering
    \small
    \resizebox{\textwidth}{!}{
    \begin{tabular}{@{}lllllr@{}}
    \toprule
    \textbf{Dataset} &
      \textbf{Languages} &
      \textbf{Label Space} &
      \textbf{Domains} &
      \textbf{Generators} &
      \textbf{Size} \\
    \midrule
    \begin{tabular}[c]{@{}l@{}}MixSet\\ \cite{mixset}\end{tabular} &
      English &
      \begin{tabular}[c]{@{}l@{}}Human-written, AI-polished\\ Human-initiated, -continued\\ AI-written, human-edited\\ Deeply-mixed text\end{tabular} &
      \begin{tabular}[c]{@{}l@{}}Email\\ News\\ Game reviews,\\ Paper abstracts,\\ Speech,\\ Blog\end{tabular} &
      \begin{tabular}[c]{@{}l@{}}GPT-4\\ Llama 2\\ Dolly\end{tabular} &
      3,600 \\
      \midrule
    \begin{tabular}[c]{@{}l@{}}DetectiAIve\\ \cite{abassy-etal-2024-llm}\end{tabular} &
      English &
      \begin{tabular}[c]{@{}l@{}}Human-written\\ AI-generated\\ Human-written, AI-polished\\ AI-written, AI-humanized\end{tabular} &
      \begin{tabular}[c]{@{}l@{}}arXiv abstracts,\\ Reddit posts,\\ Wikihow,\\ Wikipedia articles,\\ OUTFOX essays,\\ Peer reviews\end{tabular} &
      \begin{tabular}[c]{@{}l@{}}GPT-4o\\ Mistral 7B\\ Llama 3.1 8B\\ Llama 3.1 70B\\ Gemini\\ Cohere\end{tabular} &
      487,996 \\
      \midrule
    \begin{tabular}[c]{@{}l@{}}Beemo\\ \cite{beemo}\end{tabular} &
      English &
      \begin{tabular}[c]{@{}l@{}}AI-generated, AI-humanized\\ Human-written\\ AI-generated\\ AI-written, human-edited\end{tabular} &
      \begin{tabular}[c]{@{}l@{}}Generation,\\ Rewrite,\\ Open QA,\\ Summarize,\\ Closed QA\end{tabular} &
      \begin{tabular}[c]{@{}l@{}}Llama 2\\ Llama 3.1\\ GPT-4o\\ Zephyr\\ Mixtral\\ Tulu\\ Gemma\\ Mistral\end{tabular} &
      19,256 \\
      \midrule
    \begin{tabular}[c]{@{}l@{}}M4GT\\ \cite{wang-etal-2024-m4gt}\end{tabular} &
      English &
      \begin{tabular}[c]{@{}l@{}}Human-written, AI-continued\\ Human-written\\ AI-generated\end{tabular} &
      \begin{tabular}[c]{@{}l@{}}Peer review,\\ OUTFOX\end{tabular} &
      \begin{tabular}[c]{@{}l@{}}Llama 2\\ GPT-4\\ GPT-3.5\end{tabular} &
      33,912 \\
      \midrule
    \begin{tabular}[c]{@{}l@{}}Real or Fake \\ \cite{realorfake}\end{tabular} &
      English &
      \begin{tabular}[c]{@{}l@{}}Human-written\\ Human-initiated, AI-continued\end{tabular} &
      \begin{tabular}[c]{@{}l@{}}Recipes,\\ Presidential Speeches,\\ Short Stories,\\ New York Times\end{tabular} &
      \begin{tabular}[c]{@{}l@{}}GPT-2,\\ GPT-2 XL\\ CTRL\end{tabular} &
      9,148 \\
      \midrule
    \begin{tabular}[c]{@{}l@{}}RoFT-chatgpt \\ \cite{roft}\end{tabular} &
      English &
      Human-initiated, AI-continued &
      \begin{tabular}[c]{@{}l@{}}Short Stories,\\ Recipes,\\ New York Times,\\ Presidential Speeches\end{tabular} &
      GPT-3.5-turbo &
      6,940 \\
      \midrule
    \begin{tabular}[c]{@{}l@{}}Co-author\\ \cite{coauthor-dataset}\end{tabular} &
      English &
      Deeply-mixed text &
      \begin{tabular}[c]{@{}l@{}}Creative writing,\\ New York Times\end{tabular} &
      GPT-3 &
      1,447 \\
      \midrule
    \begin{tabular}[c]{@{}l@{}}TriBERT\\ \cite{tribert}\end{tabular} &
      English &
      \begin{tabular}[c]{@{}l@{}}Human-initiated, AI-continued\\ Deeply-mixed text\\ Human-written\end{tabular} &
      Essays &
      ChatGPT &
      34,272 \\
      \midrule
    \begin{tabular}[c]{@{}l@{}}LAMP\\ \cite{lamp}\end{tabular} &
      English &
      AI-generated, human-edited &
      Creative writing &
      \begin{tabular}[c]{@{}l@{}}GPT-4o\\ Claude 3.5 Sonnet\\ Llama 3.1 70B\end{tabular} &
      1,282 \\
      \midrule
    \begin{tabular}[c]{@{}l@{}}APT-Eval\\ \cite{apt-eval}\end{tabular} &
      English &
      Human-written, AI-polished &
      Based on MixSet &
      \begin{tabular}[c]{@{}l@{}}GPT-4o\\ Llama 3.1 70B\\ Llama 3.1 8B\\ Llama 2 7B\end{tabular} &
      11,700 \\
      \midrule
    \begin{tabular}[c]{@{}l@{}}HART\\ \cite{bao2025decouplingcontentexpressiontwodimensional}\end{tabular} &
      English &
      \begin{tabular}[c]{@{}l@{}}Human-written\\ Human-written, AI-polished\\ AI-generated, AI-humanized\\ AI-generated text\\ AI-generated, human-edted\end{tabular} &
       &
      \begin{tabular}[c]{@{}l@{}}GPT-3.5-turbo\\ GPT‑4o\\ Claude 3.5 Sonnet\\ Gemini 1.5 Pro\\ Llama 3.3 70B\\ Qwen 2.5 72B\end{tabular} &
      16,000 \\
      \midrule
    \begin{tabular}[c]{@{}l@{}}LLMDetect\\ \cite{cheng2024beyond}\end{tabular} &
      English &
      \begin{tabular}[c]{@{}l@{}}Human-written\\ Human-written, AI-polished\\ Human-written, AI-extended\\ AI-generated\end{tabular} &
       &
      \begin{tabular}[c]{@{}l@{}}DeepSeek v2\\ Llama 3 70B\\ Claude 3.5 Sonnet\\ GPT-4o\end{tabular} &
      64,304 \\
      \midrule
    \begin{tabular}[c]{@{}l@{}}ICNALE corpus\\ \cite{icnale}\end{tabular} &
      English &
      Human-written &
      Essays &
      \begin{tabular}[c]{@{}l@{}}Qwen 2.5\\ Llama 3.1 8B/70B\\ Llama 3.2 1B/3B\\ Mistral Small\end{tabular} &
      67,000 \\
    \bottomrule
    \end{tabular}
    }
    \caption{English fine-grained AI-generated text detection datasets overview.}
    \label{tab:cur-dataset}
\end{table*}

\begin{table}
    \centering
    \small
    \begin{tabular}{lrrr}
    \toprule
    \textbf{Subset} 
      & \multicolumn{1}{c}{\textbf{Human}}
      & \multicolumn{1}{c}{\textbf{LLM}}
      & \multicolumn{1}{c}{\textbf{Human--LLM}}\\
      & & & \multicolumn{1}{c}{\textbf{collab.}}\\
    \midrule
    Train 
      & 14,176 & 12,076 & 32,091 \\
    Validation
      & 3,038 & 2,588 & 6,876 \\
    Test
      & 3,038 & 2,588 & 6,879 \\
    \bottomrule
    \end{tabular}
    \caption{Number of examples per label in subsets in the FAIDSet dataset.}
    \label{tab:faidset-subset}
\end{table}

\begin{table}
    \centering
    \small
    \begin{tabular}{lr}
    \toprule
    \textbf{Source} & \textbf{Human Texts} \\
    \midrule
    arXiv abstracts & 2,000 \\
    VJOL abstracts & 2,195 \\
    HUST theses (English) & 4,898 \\
    HUST theses (Vietnamese) & 11,159 \\
    \bottomrule
    \end{tabular}
    \caption{Statistics of human-written text's origins in FAIDSet.}
    \label{tab:faidset-hw}
\end{table}

\section{FAIDSet Statistics and Analysis}
\label{sec:faidset-stat-analyze}

\subsection{Statistics}
\label{sec:faidset-stat}
Our FAIDSet includes 83,350 examples, which are divided into three subsets: train, validation, and test, with the ratios shown in Table \ref{tab:faidset-subset}. The dataset also comprises various sources of human-written text, as described in Table \ref{tab:faidset-hw}.

\subsection{Diverse Prompt Strategies}
\label{sec:diverse-prompt}

In order to avoid biasing our generated corpora toward some style or topic, we use a broad set of prompt templates when synthesizing LLM-generated and human--LLM collaborative texts. By varying prompt structures, content domains, and complexity, we ensure that the resulting outputs cover a wide variety of writing patterns, vocabulary, and rhetorical devices. This diversity helps our detector generalize more effectively to real-world inputs.

Concretely, we use five prompts for LLM-generated texts in Table~\ref{tab:prompt-aigen} and several categories of human--LLM collaborative texts in Tables~\ref{tab:prompt-aipolish}, \ref{tab:prompt-aicontinue}, and \ref{tab:prompt-aiparaphrase}, consecutively:
\begin{itemize}
  \item \textbf{LLM-polished:} Texts that a human initially wrote and then lightly refined by an LLM system to improve grammar, clarity, or fluency without altering the core content or intent.
  \item \textbf{LLM-continued:} Texts where a human wrote an initial portion (e.g., a sentence or paragraph), and an LLM generated a continuation that attempts to follow the original style, tone, and intent.
  \item \textbf{LLM-paraphrased:} Texts that were initially written by a human and then reworded by an LLM system to express the same meaning using different phrasing, possibly altering sentence structure or word choice while preserving the original message.
\end{itemize}

By mixing prompts across these categories, we generate a balanced corpus that mitigates over‐fitting to any one prompt pattern and better reflects the diversity of real user queries.

\begin{table*}[t!]
    \centering
    \scriptsize
    \begin{tabularx}{\textwidth}{X X}
        \toprule
        \textbf{Student Thesis} & \textbf{Paper Abstract} \\
        \midrule
        \begin{itemize}[leftmargin=0.2cm]
            \item You are a university student majoring in computer science. Please briefly summarize the main idea of the following paragraph. After that, rewrite the paragraph based on this content. Write naturally in an academic style. The rewritten paragraph should be approximately the same length in characters as the original. The original text is: \underline{\hspace{0.7cm}}
            \item In clear, structured prose, draft the section for a thesis titled \underline{\hspace{0.7cm}}, cite some related works you mentioned in the passage, and highlight the contribution. The original text is: \underline{\hspace{0.7cm}}. Please begin by briefly summarizing the main idea of the paragraph to ensure full comprehension and retention of all essential content. Then, rewrite the paragraph in a formal academic style, consistent with a university-level thesis. The rewritten section should read naturally, be coherent within the context of an academic paper, and have approximately the same character length as the original. 
            \item Assume the role of a senior software engineer. Your task is to process a paragraph from a computer science thesis using a two-step method. Firstly, you must summarize by deconstruction, which involves analyzing the original text and providing a structured summary by identifying its primary purpose, key input parameters, and main outcomes. Secondly, rewrite the paragraph based on the structured summary. The new version must be technically precise, unambiguous, and logically structured, making it easy for another engineer to understand. The original text is: \underline{\hspace{0.7cm}}. 
            \item You are a research scientist preparing a paper for a top-tier computer science conference. The original text is: \underline{\hspace{0.7cm}}. For the following paragraph from a thesis draft, you have to summarize the core contribution, which can be earned by beginning with providing a concise, one-sentence summary that captures the main scientific contribution or key finding of the paragraph. Then, rewrite the paragraph for publication by using the summary as a guide, ensuring it is written in a formal, objective, and precise tone suitable for a peer-reviewed publication. Ensure the rewritten text is information-dense yet easy for a fellow researcher to follow. 
            \item You are a university student majoring in computer science. You need to extract the main idea by writing a summary of the paragraph's main idea. Rewrite the paragraph based on the content you have summarized. The rewritten text should be in a formal academic style, read naturally, be coherent, and have approximately the same character length as the original. The original paragraph is: \underline{\hspace{0.7cm}}
        \end{itemize}
        &
        \begin{itemize}[leftmargin=0.2cm]
            \item Assume the role of a researcher with experience in writing abstracts for scientific papers. Write a short paragraph of approximately 150-200 words based on the topic conveyed by the provided file name. Start directly with the topic, presenting it clearly, objectively, and in a concise academic style. Use correct spelling and grammar, and write in a scholarly tone. Topic name: \underline{\hspace{0.7cm}}
            \item You are a computer scientist who is very familiar with abstract writing for your works, based on the title. Craft a concise \texttt{word\_count}-word abstract for a paper titled \underline{\hspace{0.7cm}}, summarizing the problem statement, methodology, key findings, and contributions. The original text: \underline{\hspace{0.7cm}}. Compose a \texttt{word\_count}-word abstract for the paper \underline{\hspace{0.7cm}}, ensuring it includes motivation, approach, results, and implications for future research. 
            \item Act like a senior researcher acting as a peer reviewer. Your task is to analyze and then rewrite the provided abstract to improve its structure and clarity. Deconstruct the abstract: First, examine the original text and break it down into four key components: What is the core problem being addressed?; What is the proposed solution or methodology?; What were the key results of the experiments?; What is the main contribution or impact of this work? After that, you must reconstruct the information from the components using only the data from your deconstructed points, and synthesize a new, cohesive abstract of approximately 150-200 words. Original abstract: \underline{\hspace{0.7cm}}.
            \item You are a research scientist specializing in the sub-field suggested by the paper's title. Your task is to generate a plausible abstract based only on the title. Based on the title, first generate a bulleted list of the likely components this paper would cover: the specific problem it probably addresses, the methodology it might propose, the kind of results one would expect, and its potential impact or contribution to the field. After that, weave these hypothesized points into a compelling and professional abstract of 150-200 words. Write it as if you were the author, confidently presenting your work. Paper title: \underline{\hspace{0.7cm}}
            \item You are a technical writer for a prestigious AI research blog. Your goal is to rewrite a standard academic abstract to make it more impactful and highlight its core breakthrough for a broader technical audience. First, read the original abstract and identify the single most important takeaway or the core breakthrough of the paper and summarize this in one sentence. Second, rewrite the abstract to be approximately 150-200 words in length. Start with a strong opening sentence that directly states the problem or the breakthrough you identified. Then, briefly explain the methodology and results, always connecting them back to why they are important. The tone should be highly professional but more engaging than a typical arXiv abstract. Preserve all technical terms and citations accurately. Paper title: \underline{\hspace{0.7cm}}. Original abstract: \underline{\hspace{0.7cm}}.
        \end{itemize}
        \\
        \bottomrule
    \end{tabularx}
    \caption{List of diverse prompt templates used to generate FAIDSet -- Label: \textbf{LLM-generated}.}
    \label{tab:prompt-aigen}
\end{table*}

\begin{table*}[t!]
    \centering
    \scriptsize
    \begin{tabularx}{\textwidth}{X X}
        \toprule
        \textbf{Student Thesis} & \textbf{Paper Abstract} \\
        \midrule
        \begin{itemize}[leftmargin=0.2cm]
            \item You are a university student majoring in computer science who has been assigned the task of polishing the paragraph below, which is excerpted from an undergraduate thesis. Improve the paragraph to make it clearer, more coherent, and more precise, while maintaining the original author's academic tone and writing style. Do not rephrase any reference materials, figure labels, or citations preserve them exactly as they appear in the original paragraph. The original text: \underline{\hspace{0.7cm}}.
            \item You are a meticulous academic editor with a specialization in computer science theses. Your task is to polish the following paragraph for conciseness and impact. Focus on eliminating redundant words and phrases, replacing weak verb constructions with stronger, more active verbs, and ensuring each sentence contributes directly and efficiently to the paragraph's central point. The core technical meaning, all specific terminology, citations, and references to figures must be preserved precisely. The original text: \underline{\hspace{0.7cm}}.
            \item Assume you are an IT student who is refining your work to make it more complete. Refine the following section excerpt for grammar, clarity, and academic style while preserving its original meaning and terminology. Provide only the polished version, without any introductory or explanatory text. The original text: \underline{\hspace{0.7cm}}.
            \item Act as a PhD candidate reviewing a section of an undergraduate thesis. Your primary goal is to enhance the logical flow and argumentative coherence of the paragraph below. Revise the text to ensure that sentences connect seamlessly with clear transitions. The paragraph should build a coherent argument from the opening sentence to the conclusion. Do not introduce new information or alter the original technical content, terminology, or references. Your focus is solely on restructuring the existing information to create a stronger, more persuasive, and logical narrative. The original text: \underline{\hspace{0.7cm}}.
            \item You are a fourth-year IT student who is refining your work to make it more complete. Enhance the academic tone, coherence, and logical flow of this thesis section without altering technical content. The original text: \underline{\hspace{0.7cm}}.
            
        \end{itemize}
        &
        \begin{itemize}[leftmargin=0.2cm]
            \item You are a researcher who has been assigned the task of polishing the paragraph below, which is excerpted from an undergraduate thesis. Improve the paragraph to make it clearer, more coherent, and more precise, while maintaining the original author's academic tone and writing style. Do not rephrase any reference materials, figure labels, or citations—preserve them exactly as they appear in the original paragraph. The original text:\underline{\hspace{0.7cm}}. 
            \item You are a scientist who is very familiar with abstract writing and refining the written abstract. Improve the coherence, precision, and formal tone of this draft abstract without introducing new content. Provide only the polished version, without any introductory or explanatory text. The original text:\underline{\hspace{0.7cm}}. 
            \item Improve the clarity and conciseness of this abstract paragraph while maintaining all original findings and terminology. Provide only the polished version, without any introductory or explanatory text. The original text: \underline{\hspace{0.7cm}}.
            \item You are a program chair for a leading academic conference, skilled at identifying impactful research. Your task is to polish the following arXiv abstract to maximize its impact and make its core contribution immediately apparent. Focus on sharpening the opening sentence to act as a compelling hook. Rephrase and reorder sentences as needed to clearly convey the main findings and highlight the significance of the work. All original terminology, data, and citations must be strictly preserved. Provide only the polished version as a single, continuous paragraph. The original text: \underline{\hspace{0.7cm}}.
            \item You are a meticulous editor for a top-tier scientific journal, such as Nature or Science. Your objective is to polish the following arXiv abstract to enhance its technical precision and information density. Scrutinize every word to ensure it is the most accurate choice. Refine phrasing to eliminate any ambiguity and, where supported by the text, replace qualitative descriptions with more specific, quantitative statements. The goal is a text where every clause delivers critical information efficiently. Do not alter the scientific findings, technical terms, or citations. The original text: \underline{\hspace{0.7cm}}.
        \end{itemize}
        \\
        \bottomrule
    \end{tabularx}
    \caption{List of diverse prompt templates used to generate FAIDSet -- Label: \textbf{LLM-polished}.}
    \label{tab:prompt-aipolish}
\end{table*}

\begin{table*}[t!]
    \centering
    \scriptsize
    \begin{tabularx}{\textwidth}{X X}
        \toprule
        \textbf{Student Thesis} & \textbf{Paper Abstract} \\
        \midrule
        \begin{itemize}[leftmargin=0.2cm]
            \item You are a university student majoring in computer science who has been assigned the task of continuing the content of the paragraph below, which is excerpted from an undergraduate thesis. Please continue the text naturally, striving to mimic the tone and writing style of the given paragraph to avoid any inconsistency in expression, while ensuring clarity and coherence in an academic style. Do not rephrase the reference materials, figure labels, or citations-preserve them exactly as they appear in the original paragraph. The original text: \underline{\hspace{0.7cm}}.
            \item You are an IT student who is writing your graduation thesis. Continue to write the section from file name \underline{\hspace{0.7cm}} thesis excerpt for approximately \texttt{word\_count} words, maintaining formal academic structure and style. Do not rephrase the reference materials, figure labels, or citations—preserve them exactly as they appear in the original paragraph. The original text: \underline{\hspace{0.7cm}}.
            \item Act like an IT student who is writing your graduation thesis. Extend the section by adding supporting detailed information for a thesis on \underline{\hspace{0.7cm}}. Do not rephrase the reference materials, figure labels, or citations—preserve them exactly as they appear in the original paragraph. The original text: \underline{\hspace{0.7cm}}.
            \item You are a computer science researcher meticulously documenting your work. Your task is to continue the paragraph starting with the sentence below. The continuation must elaborate on the underlying mechanism, process, or rationale implied by the initial statement, effectively answering the how or why. The completed paragraph should be logically sound and consistent with the topic of a thesis titled \underline{\hspace{0.7cm}}. Maintain a formal academic tone and provide only the continuation as a single, seamless paragraph. Initial sentence: \underline{\hspace{0.7cm}}.
            \item You are a final-year IT student analyzing your research findings for your graduation thesis. Continue the paragraph that begins with the key statement below by providing an analytical extension. Your writing should focus on comparing the statement to existing work, contrasting it with alternative approaches, or discussing its broader implications within the context of the thesis titled \underline{\hspace{0.7cm}}. Ensure the analysis is coherent and maintains a scholarly tone. Initial statement: \underline{\hspace{0.7cm}}.
        \end{itemize}
        &
        \begin{itemize}[leftmargin=0.2cm]
            \item You are a researcher who has been assigned the task of continuing the content of the paragraph below, which is excerpted from an undergraduate thesis. Please continue the text naturally, striving to mimic the tone and writing style of the given paragraph to avoid any inconsistency in expression, while ensuring clarity and coherence in an academic style. Do not rephrase the reference materials, figure labels, or citations-preserve them exactly as they appear in the original paragraph. The original text: \underline{\hspace{0.7cm}}
            \item You are a scientist who is very familiar with abstract writing. Add some concise concluding sentences to this partial abstract that highlight implications for future research. Do not rephrase the reference materials, figure labels, or citations-preserve them exactly as they appear in the original paragraph. The original text: \underline{\hspace{0.7cm}}.
            \item Continue the abstract by writing a closing statement that underscores the study's contributions and potential applications. Do not rephrase the reference materials, figure labels, or citations-preserve them exactly as they appear in the original paragraph. The original text: \underline{\hspace{0.7cm}}.
            \item You are a research scientist continuing the draft of a paper's abstract. The provided text introduces the core problem or context. Your task is to continue the abstract by providing a concise description of the proposed methodology or approach. Detail the key techniques, model architecture, or experimental setup used to address the problem, ensuring the description is plausible for a paper titled \underline{\hspace{0.7cm}}. The continuation must seamlessly connect to the initial text to form a single, coherent paragraph. Provide only the new text. Paper title: \underline{\hspace{0.7cm}}, initial text: \underline{\hspace{0.7cm}}.
            \item You are the lead author of a scientific paper summarizing your work. The text below already outlines the problem and methodology. Your task is to continue the abstract by presenting the key results and findings. Report on the primary outcomes, important performance metrics, or significant observations derived from your experiments. The results must be specific, quantitative where possible, and logically follow from the described method. The output must integrate smoothly with the initial text to form a single, cohesive paragraph. Initial text: \underline{\hspace{0.7cm}}.
        \end{itemize}
        \\
        \bottomrule
    \end{tabularx}
    \caption{List of diverse prompt templates used to generate FAIDSet -- Label: \textbf{LLM-continued}.}
    \label{tab:prompt-aicontinue}
\end{table*}

\begin{table*}[t!]
    \centering
    \scriptsize
    \begin{tabularx}{\textwidth}{X X}
        \toprule
        \textbf{Student Thesis} & \textbf{Paper Abstract} \\
        \midrule
        \begin{itemize}[leftmargin=0cm]
            \item You are an academic writing tutor. Your task is to perform a deep paraphrase of the following thesis excerpt. The goal is to create a version with significant structural and lexical differences from the original, while rigorously preserving the precise meaning, nuance, and all technical information. Focus on altering sentence construction and rephrasing ideas in a completely fresh way. All specific technical terms, citations, and figure labels must remain unchanged. The tone must remain formal and scholarly. The original text: \underline{\hspace{0.7cm}}.
            \item You are a Computer Science Student tasked with paraphrasing the following paragraph, which has been extracted from the thesis of an undergraduate student. Paraphrase the given thesis content while preserving its original meaning and context. Maintain clarity, coherence, and an academic tone. Do not paraphrase: References, figure labels, and citations should remain unchanged. The original text: \underline{\hspace{0.7cm}}.
            \item Paraphrase the following thesis section in a clear academic tone, preserving citations and technical terms exactly. Do not paraphrase: References, figure labels, and citations should remain unchanged. The original text: \underline{\hspace{0.7cm}}.
            \item You are a senior researcher mentoring a student on their thesis. Paraphrase the following paragraph with the primary goal of improving clarity and directness. Untangle convoluted sentences and rephrase the content using a more straightforward structure. The aim is to express the same technical information in a way that is easier for a reader to parse, without losing any nuance or academic rigor. Do not alter or rephrase technical terminology, citations, or references to figures. The original text: \underline{\hspace{0.7cm}}.
            \item Reword this thesis excerpt to improve readability and maintain its scholarly voice, keeping all references unchanged. Do not paraphrase: References, figure labels, and citations should remain unchanged. The original text: \underline{\hspace{0.7cm}}.
        
        \end{itemize}
        &
        \begin{itemize}[leftmargin=0cm]
            \item You are a researcher with paraphrasing the following paragraph, which has been extracted from the abstract of a science paper in the computer science domain. Paraphrase the given abstract content while preserving its original meaning and context. Maintain clarity, coherence, and an academic tone. Do not paraphrase: References, figure labels, and citations should remain unchanged. The original text: \underline{\hspace{0.7cm}}
            \item Rephrase this abstract in formal academic English, maintaining all original citations and technical accuracy. Do not paraphrase: References, figure labels, and citations should remain unchanged. The original text: \underline{\hspace{0.7cm}}.
            \item You are an expert scientific editor tasked with reframing an abstract to maximize its immediate impact. Perform a structural paraphrase on the following text. First, identify the core components of the abstract (Problem, Method, Results, Contribution) internally. Then, rephrase and reorder these components to lead with the main Contribution or Result, followed by the problem it solves and the method used. This inverted structure should create a fresh and compelling narrative while preserving all original information. Strict requirements: All technical terms, data, and citations must be preserved exactly. The original text: \underline{\hspace{0.7cm}}.
            \item Paraphrase this abstract paragraph to enhance clarity and flow, ensuring all technical terms and citations remain intact. Do not paraphrase: References, figure labels, and citations should remain unchanged. The original text: \underline{\hspace{0.7cm}}.
            \item You are a senior scientist adapting a specialized abstract for a broader scientific audience. Your task is to paraphrase the following text to make it more accessible to researchers in adjacent fields, without sacrificing technical rigor. Rephrase the abstract by substituting hyper-specific jargon with more widely understood technical equivalents, but only if the precise meaning is retained. The goal is for a researcher from a different subfield to quickly grasp the core concepts. Strict requirements: The abstract's original meaning, key findings, and data must be perfectly preserved. All citations and figure references must remain unchanged. The original text: \underline{\hspace{0.7cm}}
        \end{itemize}
        \\
        \bottomrule
    \end{tabularx}
    \caption{List of diverse prompt templates used to generate FAIDSet -- Label: \textbf{LLM-paraphrased}.}
    \label{tab:prompt-aiparaphrase}
\end{table*}

\section{Experimental Details and Real-World Use Cases}
\subsection{Experimental Setup}
Unless noted otherwise, we use a batch size of 64, the AdamW optimizer with a learning rate of $2 \times 10^{-5}$ and weight decay of $10^{-4}$, 50 epochs, and 2000 warm-up steps. For the fuzzy kNN component, we set top-K to 20, and we use a temperature of 0.7.

\subsection{Computational Cost}
We run all experiments on a single NVIDIA A100 (40 GB). The total wall-clock training time for FAID on FAIDSet is approximately 5 hours.

In our setup using FAISS \cite{faiss} with CPU inference, the average query latency is approximately 5ms for FAIDSet with 83k embeddings on a standard server (Intel Xeon CPU, 64GB RAM). When performing with Fuzzy KNN, it takes some small period of time to process, and thus the total average time for inference is 10ms.

\subsection{Real-world Management of Vector Database}
For FAIDSet, our entire training and validation embedding store requires approximately 200 MB, which is easily handled by standard server disks.

For larger datasets, we believe the system remains scalable:
\begin{itemize}
    \item FAISS supports disk-based indices and optimized search methods to keep memory usage low even with millions of vectors.
    \item Modern servers with hundreds of GB of disk are sufficient for storing large embedding banks, and the use of CPU-only inference (since Fuzzy KNN is GPU-free) makes the architecture cost-effective for deployment in resource-constrained environments.
\end{itemize}

\section{Analysis on Similarity across Models and Model Families}
\label{sec:analysis}

To assess the stylistic and semantic consistency of AI-generated texts, we conducted a comprehensive analysis across multiple perspectives, including N-gram distributions, text length patterns, and semantic embedding visualizations. This allowed us to study the similarities within and across model families, leading to a robust understanding of AI ``authorship'' characteristics.

\subsection{Text Distribution between LLM Families}

We analyzed the distribution of text length, measured by both word and character counts, across outputs from five LLMs: Llama-3.3-70B-Instruct-Turbo \cite{grattafiori2024llama3herdmodels}, GPT-4o-mini \cite{openai2024gpt4ocard}, Gemini~2.0, Gemini~2.0~Flash-Lite, and Gemini~1.5~Flash \cite{geminiteam2024geminifamilyhighlycapable}. Using 2,000 arXiv prompt seeds, each model generated a single output, and we plotted the resulting length distributions.

\begin{figure*}[t!]
    \centering
    \begin{subfigure}[t]{\linewidth}
        \includegraphics[width=\linewidth]{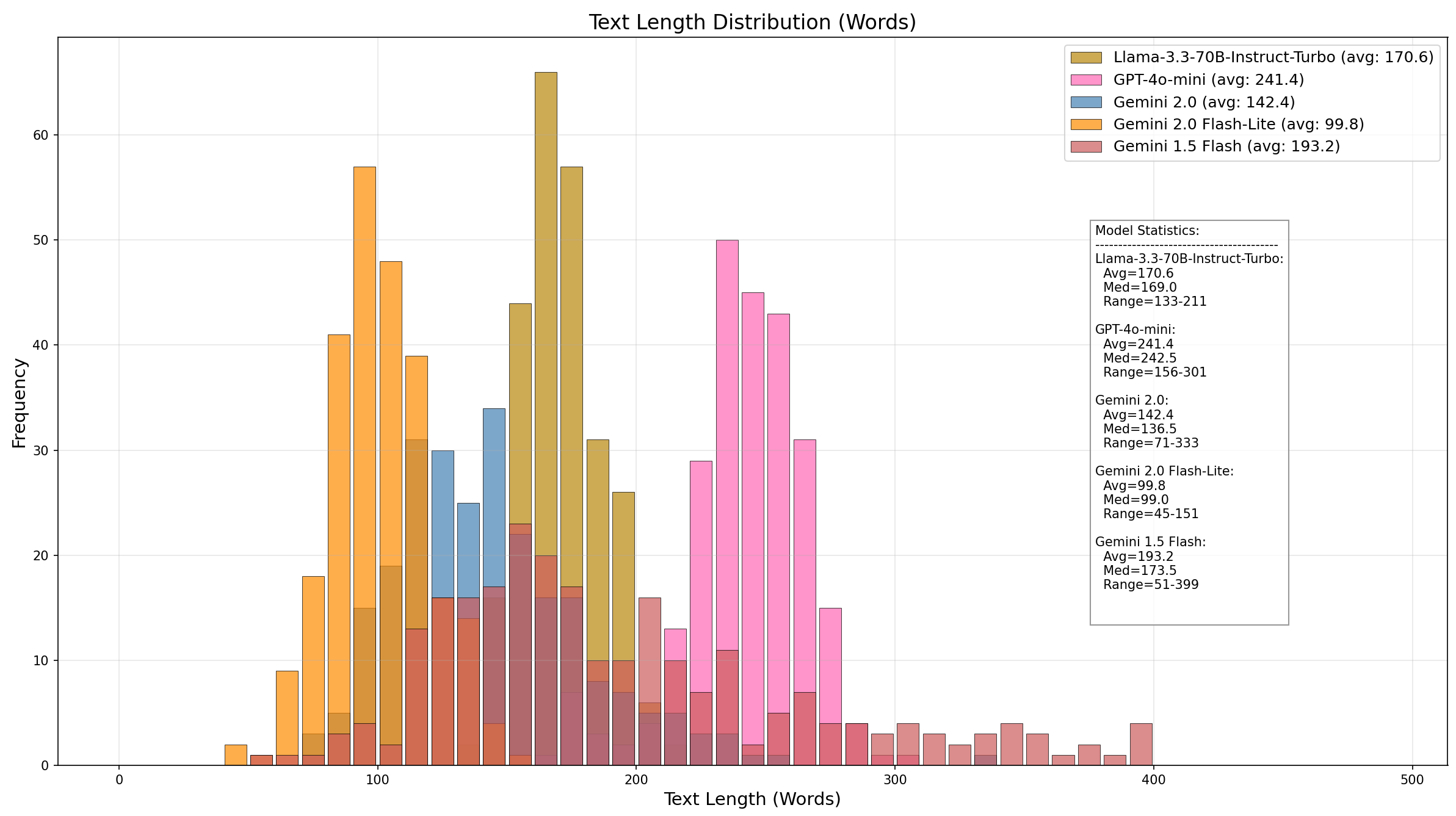}
        \caption{word length distribution across five LLMs}
        \label{fig:word_distribution}
    \end{subfigure}
    \hfill
    \begin{subfigure}[t]{\linewidth}
        \includegraphics[width=\linewidth]{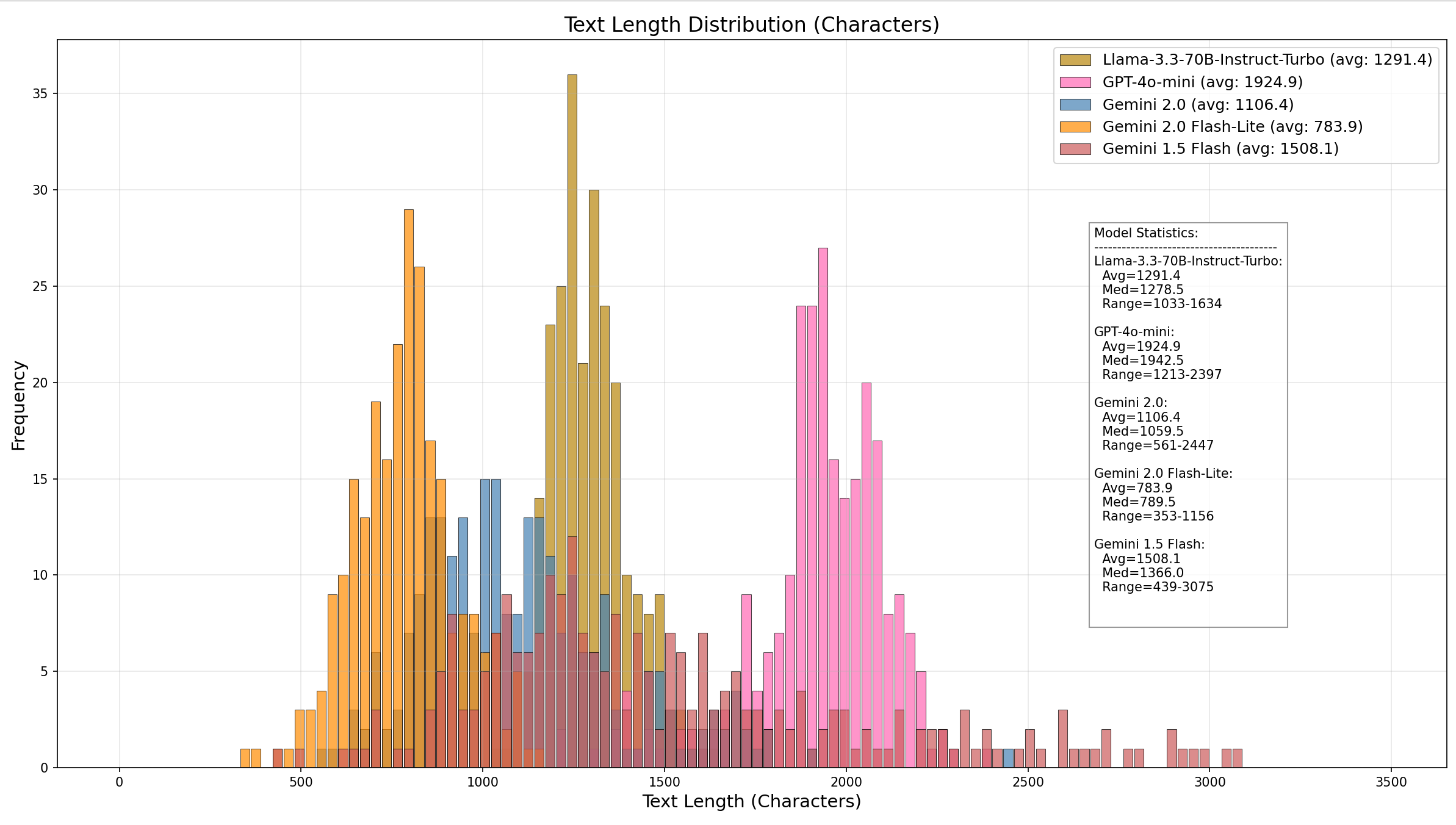}
        \caption{character length distribution across five LLMs}
        \label{fig:character_distribution}
    \end{subfigure}
    \caption{Text length distributions in words and characters across Llama-3.3, GPT-4o/4o-mini, Gemini 2.0, Gemini 2.0 Flash-Lite, and Gemini 1.5 Flash.}
    \label{fig:combined_distribution}
\end{figure*}

The results are shown in Figure~\ref{fig:combined_distribution}, where we can observe that clear family-level patterns emerge. Gemini models consistently produce shorter, more compact outputs, whereas Llama and GPT models exhibit greater variance and a stronger tendency toward longer completions. Despite differences across versions (e.g., Gemini~2.0 vs.\ Gemini~1.5~Flash), Gemini outputs remain tightly clustered in both word and character counts, indicating a shared generation strategy and strong stylistic consistency within the family. In contrast, Llama and GPT distributions show greater dispersion and inter-model variability.

This reinforces the hypothesis that text length is not only model-dependent but also family-coherent, with Gemini models forming a distinct cluster.

\subsection{Text Distribution between LLMs within the Same Family}

\begin{figure*}[t!]
    \centering
    \begin{subfigure}[t]{0.78\linewidth}
        \includegraphics[width=\linewidth]{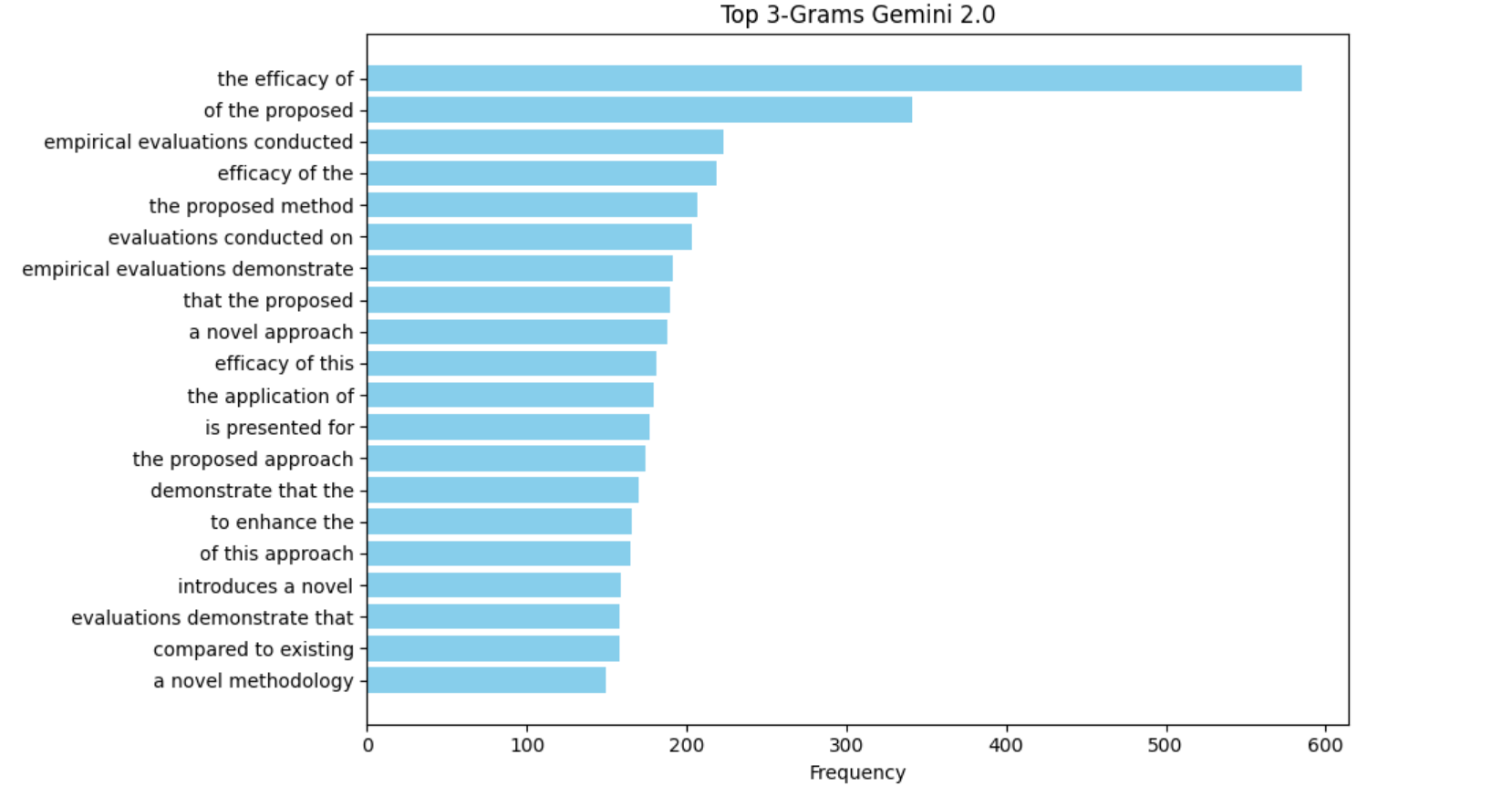}
        \caption{Top 3-grams of Gemini 2.0 (500 samples).}
        \label{fig:gemini2}
    \end{subfigure}
    \hfill
    
    \begin{subfigure}[t]{0.78\linewidth}
        \includegraphics[width=\linewidth]{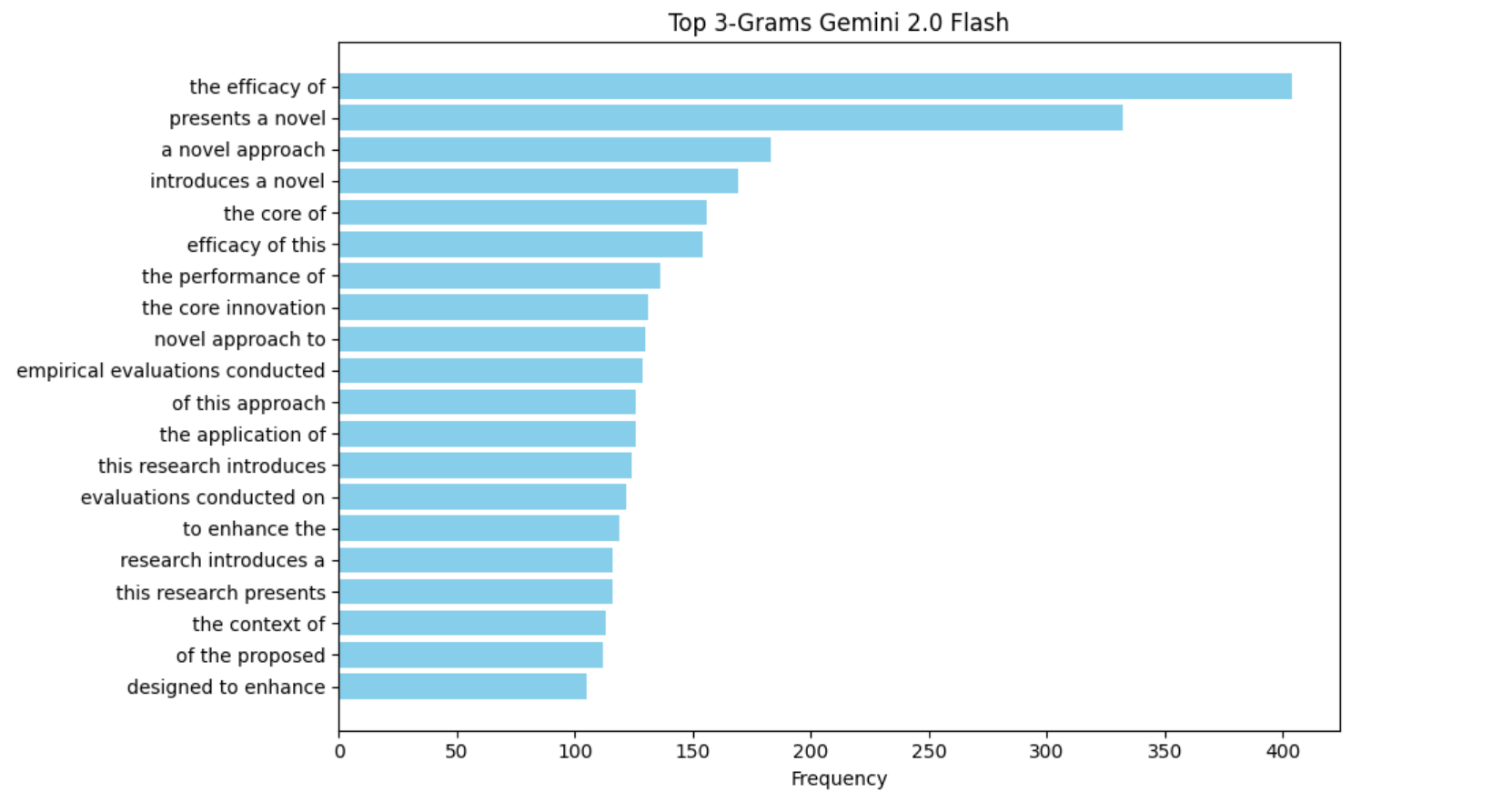}
        \caption{top 3-grams of Gemini 2.0 Flash (500 samples)}
        \label{fig:gemini2_flash}
    \end{subfigure}
    \hfill
    
    \begin{subfigure}[t]{0.78\linewidth}
        \includegraphics[width=\linewidth]{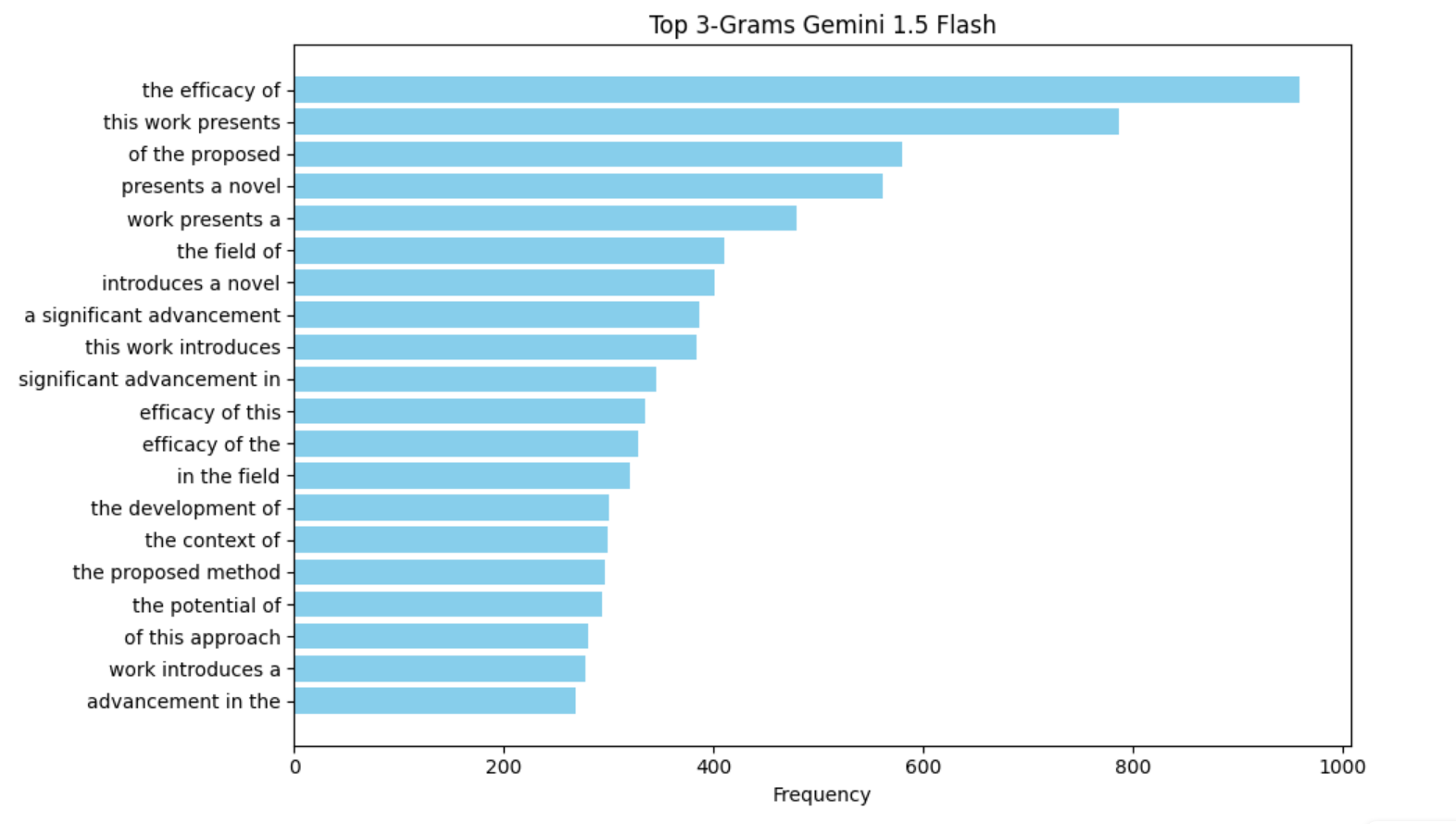}
        \caption{top 3-grams of Gemini 1.5 Flash (500 samples)}
        \label{fig:gemini15_flash}
    \end{subfigure}
    \caption{Top 20 most common trigrams from Gemini 2.0, Gemini 2.0 Flash-Lite, Gemini 1.5 Flash using 500 sample prompts.}
    \label{fig:ngram_comparison}
\end{figure*}

We performed N-gram frequency analysis on responses generated by three models within the Gemini family: Gemini 2.0, Gemini 2.0 Flash, and Gemini 1.5 Flash using 500 texts from the arXiv abstract dataset. Figure \ref{fig:ngram_comparison} highlights overlapping high-frequency tokens and similar patterns in word usage and phrase structure among the three models.

Despite minor differences in architectural speed (e.g., Flash vs. regular) or release chronology, the N-gram distributions show minimal divergence. Frequently used tokens, such as domain-specific terms and transitional phrases, appeared with nearly identical frequencies. This suggests that these models share similar decoding strategies and training biases, likely due to shared pretraining corpora and optimization techniques resulting in highly consistent stylistic patterns.

These intra-family similarities support treating model variants within a family as a unified authoring entity when performing analysis or authorship attribution.

\subsection{Embedding Visualization and Semantic Cohesion}

To explore semantic alignment in detail, we visualized the embeddings generated by an unsupervised SimCSE XLM-RoBERTa-base model on texts from two model families, Gemini and GPT, using PCA to project the high-dimensional embeddings into a lower-dimensional space for analysis.

\begin{figure*}[t!]
    \centering
    \begin{subfigure}[t]{\linewidth}
        \centering
        \includegraphics[width=0.7\linewidth]{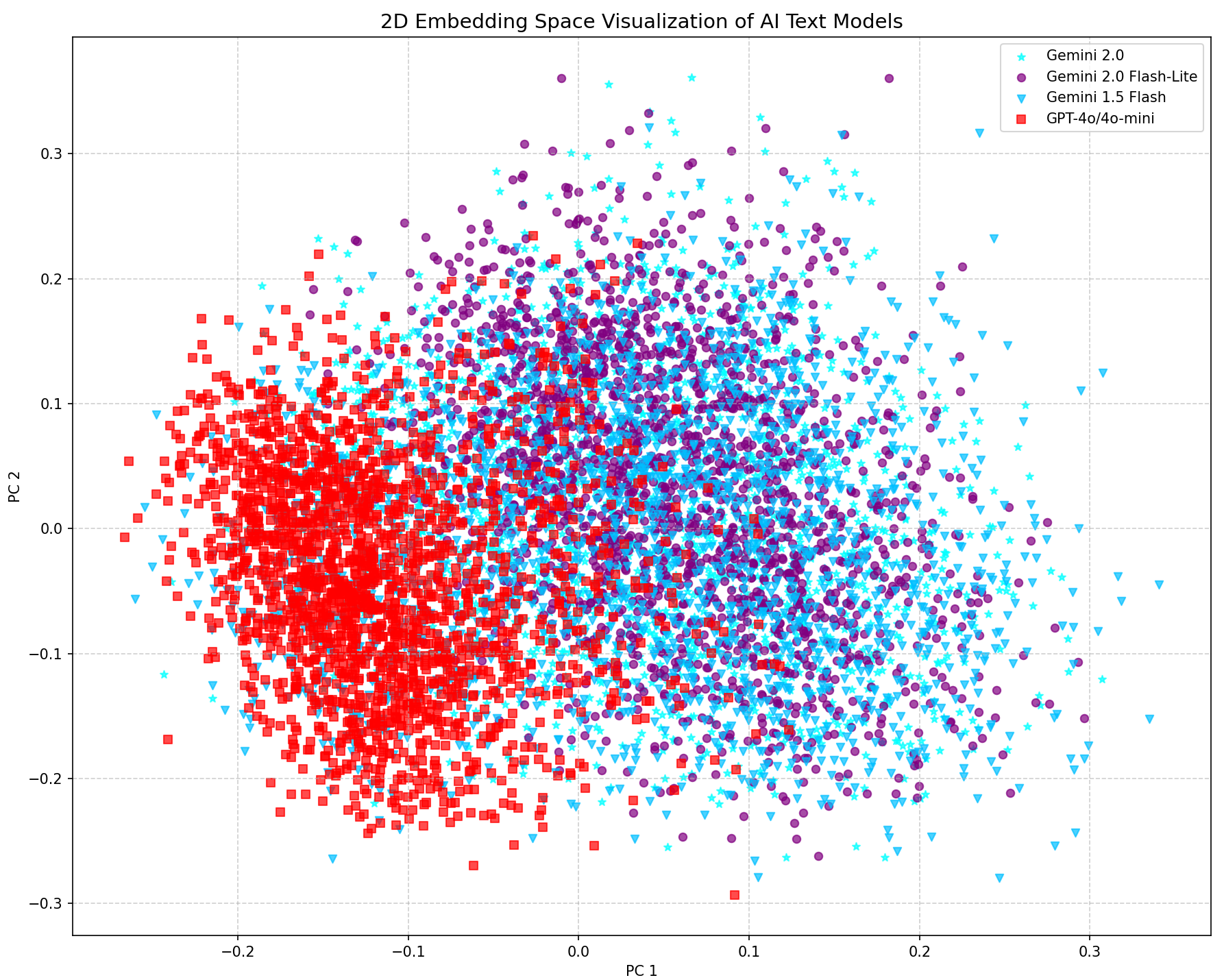}
        \caption{2D embedding space}
        \label{fig:embed2d}
    \end{subfigure}
    \hfill
    \begin{subfigure}[t]{\linewidth}
        \centering
        \includegraphics[width=0.7\linewidth]{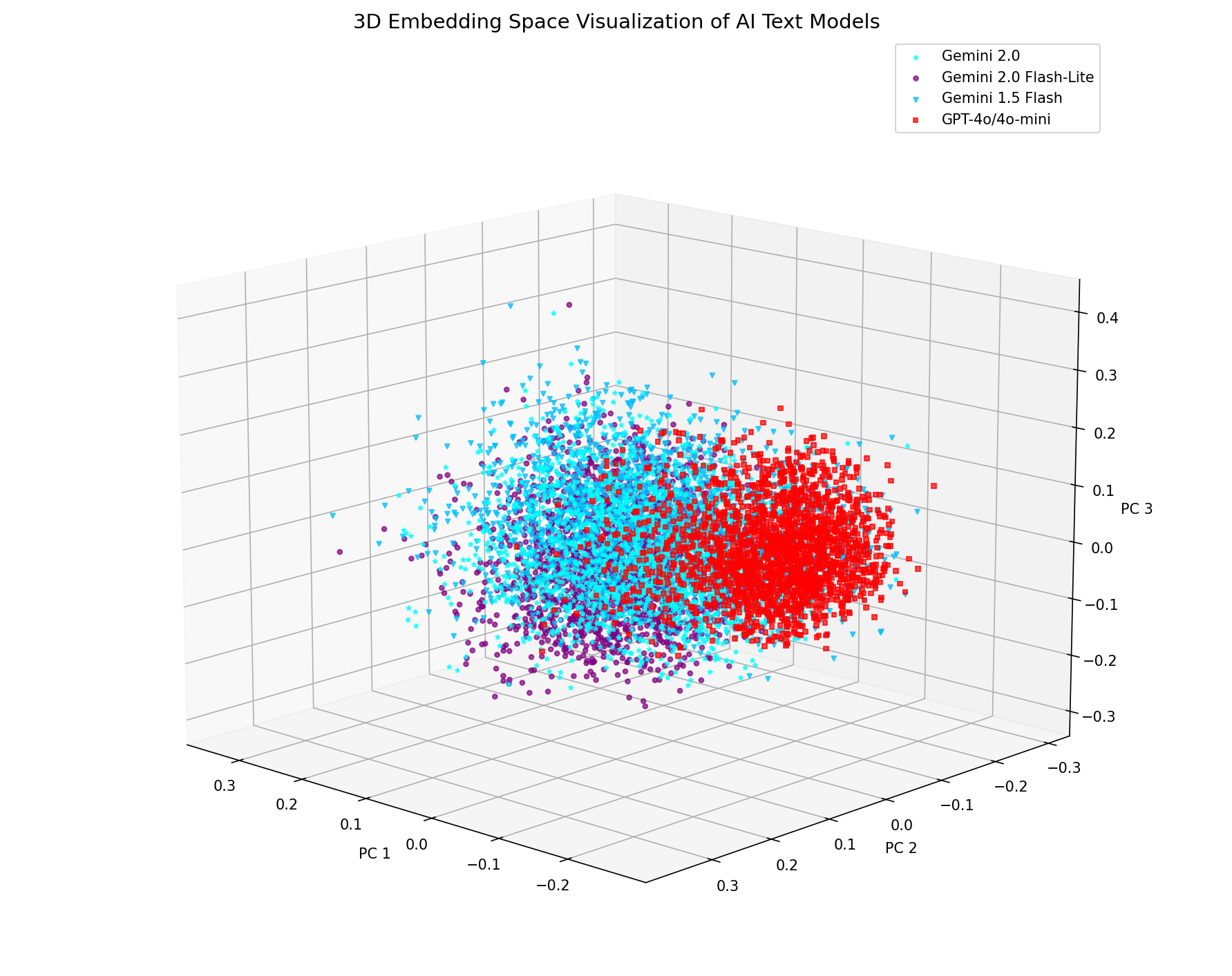}
        \caption{3D embedding space}
        \label{fig:embed_sample2k}
    \end{subfigure}
    \caption{Visualizations showing clustering behavior of Gemini model family (Gemini 2.0, Gemini 2.0 Flash-Lite, Gemini 1.5 Flash) and GPT-4o/4o-mini using 2D and 3D embeddings with sample size of 2,000 texts.}
    \label{fig:embedding_comparison}
\end{figure*}

As shown in Figure \ref{fig:embedding_comparison}, Gemini model embeddings form tight, overlapping clusters, indicating a high degree of semantic cohesion and internal consistency among their outputs. This clustering behavior remains stable across both sample sizes of 2,000 texts, suggesting that the observed patterns are not driven by sampling artifacts. In contrast, GPT-4o/4o-mini embeddings occupy a distinct region of the embedding space, exhibiting greater dispersion and noticeably less overlap with the Gemini clusters.

Overall, this visualization confirms that the Gemini family not only shares stylistic features, but also demonstrates strong semantic coherence, effectively distinguishing it from models belonging to other families at a deeper conceptual level.

\subsection{Conclusion}
The consistency observed across N-gram distributions, text length patterns, and semantic embeddings among Gemini models substantiates our decision to treat each LLM family as an author. These models demonstrate coherent writing styles, shared lexical preferences, and tightly clustered semantic representations, hallmarks of unified authorship. Conversely, inter-family comparisons show clear separability, emphasizing the distinctiveness of each LLM family's writing behavior.

\section{Model Selection for Detector}
\label{sec:model-select}

\begin{table*}[t!]
    \centering
    \resizebox{\textwidth}{!}{
    \begin{tabular}{lccccccccc}
    \toprule
    \multirow{2}{*}{\textbf{Model}} & \multirow{2}{*}{\textbf{\#. Params}} 
      & \multicolumn{4}{c}{\textbf{Known Generators}} 
      & \multicolumn{4}{c}{\textbf{Unseen Generators}} \\
      & & \textbf{Acc $\uparrow$} & \textbf{F1-macro $\uparrow$} & \textbf{MSE $\downarrow$} & \textbf{MAE $\downarrow$} 
      & \textbf{Acc $\uparrow$} & \textbf{F1-macro $\uparrow$} & \textbf{MSE $\downarrow$} & \textbf{MAE $\downarrow$} \\
    \midrule
    RoBERTa-base & 125M
      & 80.09 & 76.22 & 0.7328 & 0.3778 
      & 73.45 & 69.10 & 0.8901 & 0.4320 \\
    FLAN-T5-base & 248M
      & 80.19 & 75.77 & 0.7783 & 0.3947 
      & 72.80 & 68.55 & 0.9123 & 0.4467 \\
    e5-base-v2 & 109M
      & 81.53 & 77.90 & 0.8023 & 0.4086 
      & 74.21 & 70.15 & 0.8804 & 0.4392 \\
    Multilingual-e5-base & 278M
      & 91.41 & 90.82 & 0.3436 & 0.1732 
      & 85.32 & 84.50 & 0.5102 & 0.2543 \\
    XLM-RoBERTa-base & 279M
      & 91.90 & 90.63 & 0.2345 & 0.1190 
      & 86.75 & 85.20 & 0.4125 & 0.2104 \\
    \midrule
    Sup-SimCSE-RoBERTa-base & 279M
      & 81.22 & 78.88 & 0.7102 & 0.3619 
      & 74.00 & 71.30 & 0.8420 & 0.4251 \\
    UnSup-SimCSE-RoBERTa-base & 279M
      & 82.19 & 79.38 & 0.7156 & 0.3637 
      & 75.10 & 72.40 & 0.8305 & 0.4207 \\
    UnSup-SimCSE-XLM-RoBERTa-base & 279M
      & \textbf{92.12} & \textbf{91.75} & \textbf{0.1904} & \textbf{0.0958} 
      & \textbf{87.45} & \textbf{86.90} & \textbf{0.3507} & \textbf{0.1802} \\
    \bottomrule
    \end{tabular}
    }
    \caption{Model selection on known vs.\ unseen generators. The best results in each column are in \textbf{bold}.}
    \label{tab:model-select}
\end{table*}

In order to identify the best encoder for our classification task, we evaluated each candidate model on both known generators (in the FAIDSet test data) and the new unseen-generators test set, which is introduced in Section~\ref{sec:dataset}.

Each transformer-based encoder was fine-tuned on FAIDSet training data and then used to predict labels on the two evaluation splits. Table \ref{tab:model-select} summarizes the accuracy, F1-macro, Mean Squared Error (MSE), and Mean Absolute Error (MAE) for each model under both conditions. 

\paragraph{Base model comparison.} We first evaluated three popular monolingual models: RoBERTa‑base, Flan‑T5‑base, and e5‑base‑v2 using the same training and evaluation splits. RoBERTa‑base \cite{roberta} and e5‑base‑v2 \cite{e5base} achieved the most balanced trade‑off between classification accuracy and regression error (MSE, MAE), while Flan‑T5‑base \cite{flant5} lagged slightly in F1‑macro. These results indicate that a stronger encoder backbone yields more robust performance, motivating the exploration of multilingual variants for further gains.

\paragraph{Multilingual variants.} We next evaluated XLM\text{-}RoBERTa\text{-}base \cite{xlmroberta} and Multilingual\text{-}e5\text{-}base \cite{multie5base}. Both models benefit from cross-lingual pretraining, which in our setting improves the representation of the diverse linguistic patterns present in FAIDSet. Notably, XLM\text{-}RoBERTa\text{-}base yields a substantial improvement across all evaluation metrics, indicating that its multilingual training enhances generalization even when applied to predominantly monolingual inputs.

\paragraph{Contrastive learning with SimCSE.}
Finally, we incorporated contrastive learning via SimCSE \cite{princetonsimcse} to refine sentence embeddings. We evaluated supervised (trained on NLI data) and unsupervised (trained on the Wikipedia corpus) SimCSE variants applied to RoBERTa\text{-}base. The unsupervised variant outperformed its supervised counterpart, aligning with prior findings that unsupervised SimCSE produces stronger semantic encoders. Based on these results, we applied unsupervised SimCSE to XLM\text{-}RoBERTa\text{-}base \cite{ethroberta}, achieving the highest accuracy and lowest error rates.

Based on these experiments, we selected the \textbf{unsupervised SimCSE XLM‑RoBERTa‑base} model for our final system.

\begin{table*}[t!]
    \centering
    \small
    \begin{tabular}{lcccc}
    \toprule
    \multirow{2}{*}{\textbf{Algorithm}} 
      & \multicolumn{2}{c}{\textbf{Known Generators}} 
      & \multicolumn{2}{c}{\textbf{Unseen Generators}} \\
      & \textbf{Accuracy $\uparrow$} & \textbf{F1-macro $\uparrow$} 
      & \textbf{Accuracy $\uparrow$} & \textbf{F1-macro $\uparrow$} \\
    \midrule
    k-Nearest Neighbors (KNN) 
      & 90.52 & 90.21 & 85.37 & 84.95 \\
    k-Means 
      & 88.13 & 87.48 & 80.22 & 79.81 \\
    Fuzzy k-Nearest Neighbors
      & \textbf{95.18} & \textbf{95.05} & \textbf{93.31} & \textbf{93.25} \\
    Fuzzy C-Means 
      & 92.67 & 92.31 & 90.04 & 89.53 \\
    \bottomrule
    \end{tabular}
    \caption{Comparison of clustering algorithms on known vs.\ unseen generators. The best results are shown in \textbf{bold}.}
    \label{tab:cluster-select}
\end{table*}

\section{Ablation Study}
\label{sec:ablation-study}

\subsection{The Need to Use a Vector Database}
\label{subsec:vectordb}
We first applied the trained model to classify the input text and observed a substantial performance drop of 15–30\% when evaluated on unseen data. This degradation underscores the model's limited ability to generalize beyond its training distribution and its sensitivity to distributional shifts. To address this issue, we decided to integrate a vector database that stores dense embeddings of all examples, including both labeled instances and unlabeled data encountered during inference.
By indexing and retrieving semantically similar examples during inference, the vector database serves as a flexible, scalable memory module that helps bridge the gap between the training and test distributions. This retrieval-based mechanism enhances the classifier's robustness to domain shifts and unseen generators by grounding predictions in stylistically and semantically related examples.

Specifically:
\begin{itemize}
    \item \textbf{Robust Domain Adaptation:} New inputs are matched against a broad, continuously growing repository of embeddings, allowing the classifier to leverage analogous instances from related domains without full retraining.
    \item \textbf{Generator-Independent Coverage:} As novel text generators emerge, their embeddings populate the database; retrieval naturally adapts to new styles or patterns by finding the closest existing vectors.
\end{itemize}

\subsection{Clustering Algorithm Selection}
\label{subsec:cluster-select}

To improve our detector's robustness against unseen domains and generators, we evaluated four clustering strategies in our vector database. Each algorithm was tasked with grouping text samples into human‐written, AI‐generated, and human--LLM collaborative categories, using both known‐generator data (held out from training) and entirely unseen generator data. For evaluation, we used accuracy and F1-macro score.

We encoded each example using the penultimate layer of our classification model, then applied clustering within the vector database to assign soft or hard cluster labels corresponding to the three classes. The results are shown in Table \ref{tab:cluster-select}, where we can see:

\begin{itemize}
  \item \textbf{Traditional algorithms} show reasonable performance on held‐out known generators, but degrade notably on unseen generators.
  \item \textbf{Fuzzy C‐Means} leverages membership degrees to handle overlapping distributions, improving both measures by 4\% over k‐Means, with smaller degradation on unseen data.
  \item \textbf{Fuzzy KNN} combines local neighbor information with fuzzy membership, achieving the best overall performance.
\end{itemize}

Given its superior ability to adapt to novel domains and generators through weighted neighbor voting and soft cluster assignments, we adopt \textbf{Fuzzy k-Nearest Neighbors} as the clustering component in our overall architecture.

\end{document}